\documentclass{article}

\usepackage{longcat_style}
\usepackage{adjustbox}
\usepackage[utf8]{inputenc} %
\usepackage[T1]{fontenc}    %
\usepackage{newunicodechar}
\usepackage{hyperref}       %
\usepackage{xcolor}
\usepackage[normalem]{ulem} %
\hypersetup{
    colorlinks=true,      %
    linkcolor=blue,      %
    urlcolor=blue,       %
    citecolor=blue,      %
    linkbordercolor=blue, %
    urlbordercolor=blue,
    citebordercolor=blue,
    pdfborderstyle={/S/U/W 1}, %
}
\usepackage{float}
\usepackage{url}            %
\usepackage{booktabs}       %
\usepackage{amsfonts}       %
\usepackage{nicefrac}       %
\usepackage{microtype}      %
\usepackage{lipsum}		%
\usepackage{graphicx}
\usepackage{natbib}
\usepackage{doi}
\usepackage{amsmath}
\usepackage{amssymb} %
\usepackage{xspace}
\usepackage{enumitem}
\usepackage{multirow}
\usepackage{subcaption} 
\usepackage{makecell}
\usepackage{hyperref, cleveref}
\usepackage{pifont}
\usepackage[inkscapelatex=false]{svg}
\usepackage{subcaption}
\usepackage{CJKutf8} % add by wjn
\usepackage{tabularx} % add by wjn

\setlist[itemize]{leftmargin=*}
\setlist[enumerate]{leftmargin=*}
\setlist[description]{leftmargin=*}

\usepackage{colortbl}
\definecolor{mygray}{gray}{.88}
\definecolor{mycyan}{cmyk}{.15,0,0,0}
\definecolor{mycyan2}{cmyk}{.85,0,0,0}

\definecolor{mygreen}{rgb}{0.19, 0.79, 0.02}
\definecolor{midnightgreen}{rgb}{0.0, 0.29, 0.33}

\newcommand{\longcat}{LongCat-Flash-Thinking-2601\xspace}

\definecolor{midnightgreen}{rgb}{0.0, 0.29, 0.33}

\title{\longcat Technical Report}

\author{ Meituan LongCat Team \\
	\texttt{longcat-team@meituan.com} \\
}

\renewcommand{\headeright}{\raisebox{-0.2\height}{\includegraphics[height=1.8em]{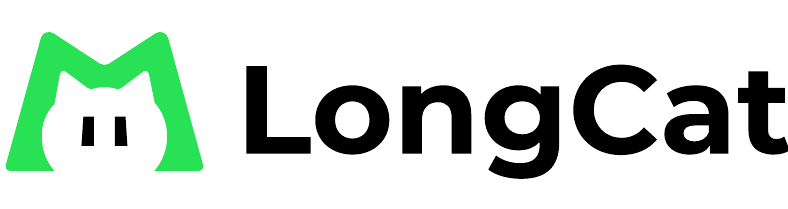}}}
\renewcommand{\shorttitle}{\longcat Technical Report}

\begin{document}
\maketitle

\begin{abstract}

We introduce \longcat, a 560-billion-parameter open-source Mixture-of-Experts (MoE) reasoning model with superior agentic reasoning capability.
\longcat achieves state-of-the-art performance among open-source models on a wide range of agentic benchmarks, including agentic search, agentic tool use, and tool-integrated reasoning.
Beyond benchmark performance, the model demonstrates strong generalization to complex tool interactions and robust behavior under noisy real-world environments.
Its advanced capability stems from a unified training framework that combines domain-parallel expert training with subsequent fusion, together with an end-to-end co-design of data construction, environments, algorithms, and infrastructure spanning from pre-training to post-training.
In particular, the model’s strong generalization capability in complex tool-use are driven by our in-depth exploration of environment scaling and principled task construction.
To optimize long-tailed, skewed generation and multi-turn agentic interactions, and to enable stable training across over 10,000 environments spanning more than 20 domains, we systematically extend our asynchronous reinforcement learning framework, DORA, for stable and efficient large-scale multi-environment training.
Furthermore, recognizing that real-world tasks are inherently noisy, we conduct a systematic analysis and decomposition of real-world noise patterns, and design targeted training procedures to explicitly incorporate such imperfections into the training process, resulting in improved robustness for real-world applications.
To further enhance performance on complex reasoning tasks, we introduce a Heavy Thinking mode that enables effective test-time scaling by jointly expanding reasoning depth and width through intensive parallel thinking.
We release our checkpoints to facilitate future research and real-world applications of agentic systems.

\textbf{LongCat Chat}: \href{https://longcat.ai}{https://longcat.ai} \\
\textbf{Huggingface}: \href{https://huggingface.co/meituan-longcat/LongCat-Flash-Thinking-2601}{https://huggingface.co/meituan-longcat/LongCat-Flash-Thinking-2601} \\
\textbf{Github}: \href{https://github.com/meituan-longcat/LongCat-Flash-Thinking-2601}{https://github.com/meituan-longcat/LongCat-Flash-Thinking-2601} \\

\end{abstract}

\begin{figure}[!htb]
    \centering
    \includegraphics[width=0.85\textwidth]{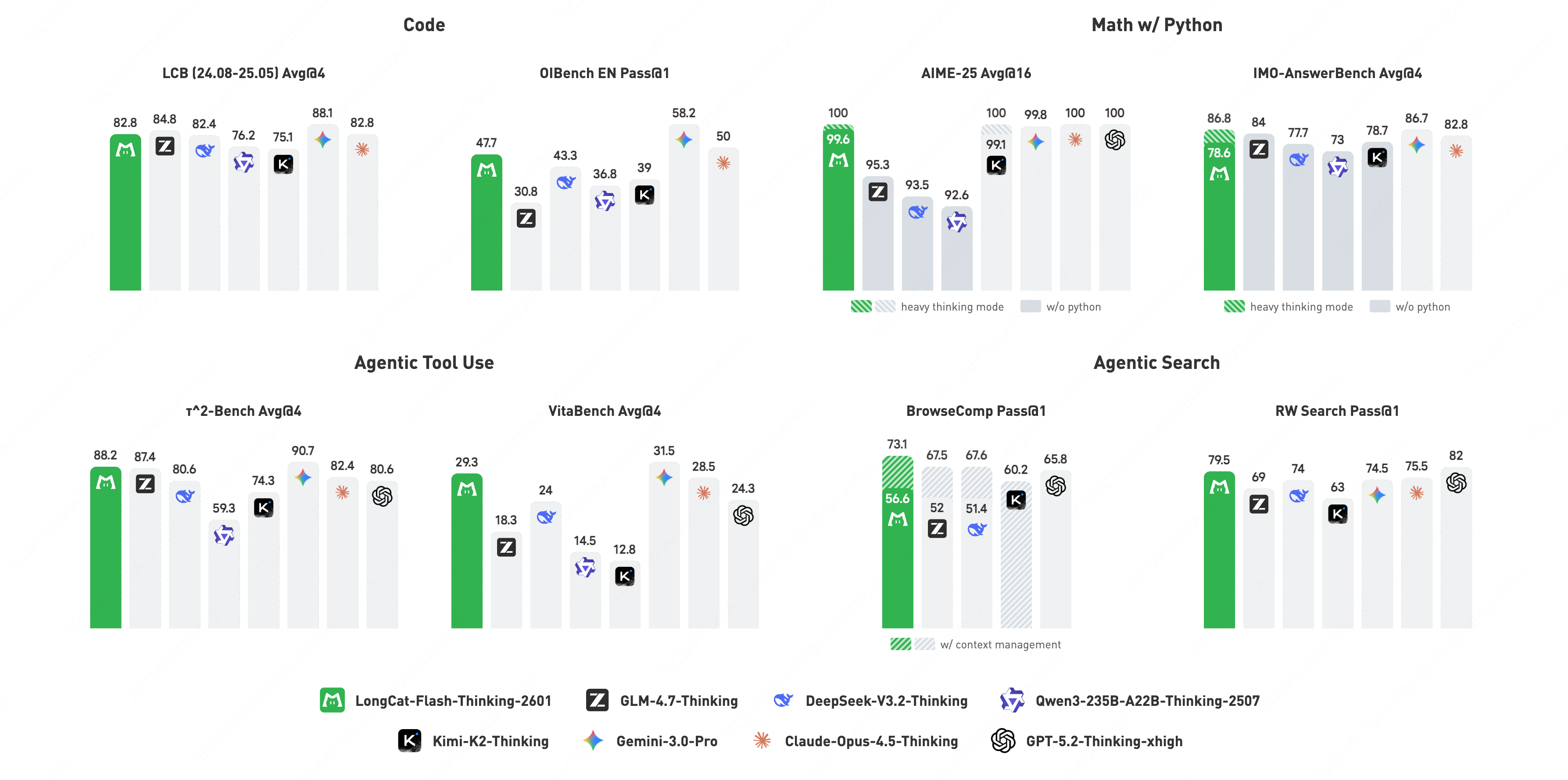}
    \caption{Benchmark performance of \longcat.}
\label{fig: benchmark_overview}
\end{figure}

\section{Introduction}
\label{sec:intro}

Recent advances in reasoning models have led to rapid progress on complex tasks such as mathematics and programming, in some cases even surpassing top human experts~\citep{Deepseek-R1, deepseek-math-v2, Claude-Opus-4.5, Gemini3}.
The natural next question then arises: how can such complex problem-solving capabilities be applied to solve complex real-life tasks, and how can complex problem-solving capabilities be extended beyond this point?
As intrinsic reasoning ability approaches its limits, we identify that interaction with external environments emerges as a key mechanism for further progress~\citep{Kimi_K2_web_doc, DeepSeek-V3.2}.
From this perspective, agentic reasoning can be understood as the ability to solve complex problems through adaptive interaction with external environments.
Beyond internal deliberation, advanced agentic reasoning capability requires models to determine when and how to interact with the environment, and to effectively integrate environmental feedback to sustain and advance the reasoning process.
In this way, reasoning and interaction naturally interleave and reinforce each other, jointly enabling more general and powerful problem-solving behavior~\citep{MiniMax-M2}.
However, enabling such agentic reasoning capability poses substantial challenges for existing models and training pipelines.
Agentic tasks typically involve long-horizon trajectories, heterogeneous environments, and long-tailed interaction dynamics, which place new demands on data curation, environment construction, reinforcement learning strategies, and system-level infrastructure spanning from pre-training to post-training stage.

In this work, we introduce \longcat, a powerful and efficient Mixture-of-Experts (MoE) reasoning model with 560B total parameters and 27B activated parameters on average per token, featuring strong agentic reasoning capability.
The pre-training of \longcat largely follows the recipe of LongCat-Flash-Chat~\citep{longcat-flash}, retaining the original data distribution to preserve competitive general reasoning performance.
Building on this foundation, we further extend the model toward large-scale agentic reasoning through a carefully designed mid-training stage.
Compared to traditional reasoning, agentic behaviors typically involve long-horizon trajectories with proactive tool invocations.
However, such interaction patterns are extremely scarce in real-world corpora, where most data consist primarily of natural language.
As a result, a raw model is largely unfamiliar with agentic interaction dynamics, leading to inefficient exploration during the reinforcement learning stage.
To address this challenge, we expose the model to moderate-scale synthesized structured agentic trajectories during mid-training, providing a strong initialization for agentic reasoning behaviors.

In post-training, we focus on efficiently scaling RL compute to improve agentic reasoning capabilities.
Beyond specific agentic scenarios such as search and code, we further identify generalization across tasks and environments as a core characteristic of advanced agentic behavior.
Such capability is primarily acquired through interaction with diverse environments, which therefore serve as the learning playground.
To this end, we design an automated environment scaling pipeline to construct complex and diverse environments covering over 20 domains, while strictly preserving executability and verifiability.
As the resulting environments exhibit substantial heterogeneity in both domain and difficulty, training over them requires a careful co-design of training strategies and infrastructure support to ensure stable and effective multi-domain environment learning.
To support large-scale agentic reinforcement learning under this setting, we extend our multi-version asynchronous training system, Dynamic ORchestration for Asynchronous Rollout (DORA), to enable scalable and stable training with up to 32,000 environments executing concurrently.
However, generalization to real-world settings remains highly challenging, as real-world environments are inherently imperfect.
To bridge the gap between idealized training settings and real-world deployment, we systematically analyze real-world noise and design an automated pipeline to progressively incorporate multi-type and multi-level environmental imperfections into the multi-domain environment training process.

To further push reasoning capability beyond existing limits, \longcat incorporates a Heavy Thinking Mode that enables effective test-time scaling of reasoning.
This mode decomposes challenging problem solving into complementary stages that jointly expand both reasoning width and depth, allowing the model to explore diverse solution paths while progressively refining its reasoning.
An additional reinforcement learning stage is introduced to strengthen the model’s ability to aggregate and refine intermediate reasoning outcomes, further enhancing the effectiveness.

While retaining strong competitiveness on general reasoning benchmarks, \longcat achieves state-of-the-art performance among open-source models across a wide range of agentic benchmarks, while demonstrating strong generalization to out-of-distribution real-world agentic scenarios.
Notably, \longcat attains 73.1\% on BrowseComp, 77.7\% on RWSearch, 88.2\% on $\tau^2$-Bench, and 29.3\% on VitaBench, establishing it as the leading open-source model for agentic search and agentic tool-use tasks.

Our work presents three core contributions:
\begin{itemize}
    \item \textbf{Environment Scaling and Multi-Domain Environment Training.}
    We develop a scalable environment construction and task generation framework that produces a large collection of high-quality, executable, and verifiable agentic environments.
    Building on this, we extend our asynchronous reinforcement learning infrastructure to support stable and efficient multi-domain environment training, enabling the acquisition of generalizable agentic skills across diverse domains.

    \item \textbf{Robust Agentic Training under Noisy Environments.}
    To address the inherent imperfection of real-world environments, we systematically analyze major sources of environmental noise and design an automated pipeline to inject multi-type and multi-level noise into training environments.
    A curriculum-based reinforcement learning strategy is adopted to progressively increase noise complexity, substantially improving robustness and performance under imperfect conditions.

    \item \textbf{Heavy Thinking Mode for Test-Time Scaling.}
    We introduce a Heavy Thinking Mode that enables effective test-time scaling of reasoning by jointly expanding reasoning width and depth.
    Through parallel trajectory exploration and iterative reasoning refinement, this mode further enhances performance on challenging reasoning and agentic tasks.
\end{itemize}

\section{Pre-Training}
\label{sec:mid_training}
Our model builds upon the pretraining recipe of LongCat-Flash-Chat~\citep{longcat-flash}, inheriting its data distribution to preserve strong general-purpose language and reasoning capabilities.
Beyond traditional reasoning, agentic behaviors typically involve long-horizon trajectories with proactive tool invocations.
This brings two additional challenges: (i) substantially increased demands on long-context modeling efficiency, and (ii) the scarcity of large-scale agentic trajectories in the real world.
To address the long-context requirements of agentic reasoning, we adopt a staged mid-training procedure with progressively increasing context lengths, allocating 500B tokens to the 32K/128K stages and an additional 40B tokens to the 256K stage.
Since large-scale reinforcement learning alone is often inefficient and unstable without a model that has already been primed with basic agentic behaviors, we choose to expose the model to moderate-scale agentic data at this stage.
To mitigate the scarcity of real-world agentic trajectories, we construct a hybrid data synthesis pipeline to collect and construct agentic training data.
Furthermore, we predict optimal hyperparameters in the mid-training stage based on any pre-trained checkpoint, specifically designed to minimize the computational cost of finding the best configuration (see Appendix \ref{sec:optimal_hyperparameter_prediction}).

\begin{figure}[t]
    \centering
    \includegraphics[width=0.95\textwidth]{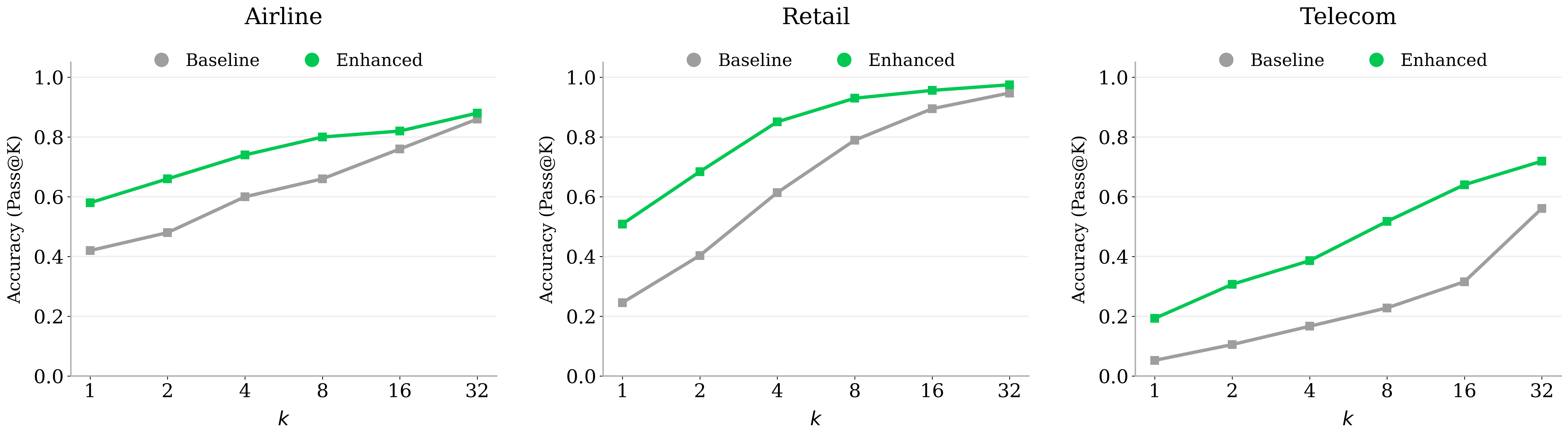}
    \caption{The comparison of agentic capability (\textit{pass@k}. ``Baseline'' represents the model undergoing the old mid-training recipe, while ``Enhanced`` represents the model undergoing the agentic-enhanced mid-training recipe.}
\label{fig:midtrain_pass_at_k}
\end{figure}

Following the original mid-training recipe to preserve general reasoning capability~\citep{zhang2025expanding,tu2025survey}, we further augment the data distribution with structured agentic trajectories~\citep{xu2026unlocking}.
Since large-scale, high-quality agentic data—especially long-horizon trajectories involving reasoning, planning, and interaction—is extremely scarce, we construct a hybrid data synthesis framework to fill this gap.
Specifically, our framework draws from two complementary sources: unstructured text and executable environments, corresponding to \textit{text-driven synthesis} and \textit{environment-grounded synthesis}, respectively.
The former provides broad semantic and task diversity, while the latter ensures logical consistency and executability.
In addition, to explicitly strengthen planning ability—a core component of agentic reasoning that is difficult to acquire from existing data—we further design a dedicated planning-centric data construction strategy. As shown in Figure \ref{fig:midtrain_pass_at_k}, the model trained on this enhanced mid-training recipe demonstrates superior agentic capability boundaries as demonstrated by larger \textit{pass@k} on the $\tau^2$-Bench.

\paragraph{Text-driven synthesis}
Large-scale text corpora contain abundant implicit procedural knowledge, such as tutorials, instructions, and multi-step problem-solving workflows.
We leverage this property to mine and restructure such latent procedures into explicit agentic interaction trajectories using the following pipeline:
\begin{itemize}
    \item \textbf{Text Filtering \& Tool Extraction}:  
    We strategically identify text segments that exhibit rich, multi-step workflows. From these segments, we define potential functions and extract corresponding function-call lists, converting implicit procedures into explicit tool schemas.
    
    \item \textbf{Synthesis \& Refinement}:  
    We translate abstract workflows and raw text into concrete, multi-turn user-agent interactions. To ensure the robustness and breadth of the dataset, we apply extensive agentic pattern diversity enhancement and rigorous quality filtering, ensuring the final trajectories cover a diverse array of interaction scenarios and task domains.
\end{itemize}
To further increase the structural complexity of the synthesized trajectories, we apply two decomposition-based augmentations:

\begin{itemize}
    \item \textbf{Tool Decomposition}:  
    Starting from simple tool invocation trajectories, we iteratively hide parts of the tool parameters into the environment. 
    Correspondingly, we synthesize model interactions to extract these parameters, progressively building more complex trajectories.
    
    \item \textbf{Reasoning Decomposition}:  
    For each action step in the model’s output, we generate multiple alternative candidates and replace the original action with these alternatives. 
    We then synthesize the model reasoning step to select the most appropriate candidate, transforming the trajectory into a decision-making process that reflect the agent’s reasoning across multiple steps.
\end{itemize}

\paragraph{Environment-grounded synthesis} In addition to text-derived trajectories, we also construct agentic data directly from executable environments to guarantee logical correctness and execution consistency.
We implement lightweight Python environments for collected toolsets and generate trajectories through controlled tool-chain sampling and execution verification:
\begin{itemize}
    \item \textbf{Environment Construction \& Dependency Modeling}:
    Based on existing tool definitions, we implement lightweight, verifiable Python environments. We explicitly model the logical dependencies between tools, constructing a directed graph where nodes represent tools and edges represent parameter dependencies. This structure allows us to sample tool invocation chains of varying complexity and depth systematically.
    
    \item \textbf{Reverse-Synthesis \& Execution Verification}:
    We sample valid tool execution paths from the dependency graph and employ a reverse-engineering approach to synthesize corresponding system prompt for user simulators that align with the chosen tool chain. Crucially, the correctness of each trajectory is validated by executing the code and verifying the final state of the environment database. This ensures that the synthesized data is grounded in actual execution logic.
\end{itemize}

\paragraph{Planning-Oriented Data Augmentation}
We observe that agentic reasoning critically depends on planning ability, which governs how the model decomposes complex goals, explores alternatives, and commits to intermediate decisions. 
However, such planning-centric behaviors are poorly represented in existing data and are difficult to acquire at scale.
To explicitly strengthen this capability during mid-training, we design a targeted data construction strategy that transforms existing trajectories into planning-centric decision-making processes.
The first type of data focuses on synthesizing effective problem decomposition trajectories paired with correct initial action selections, providing supervision for coarse-to-fine planning and early-stage decision making.
The second type starts from complete interaction trajectories and further enriches them by generating multiple alternative candidates at each decision step. The model is then trained to reason over and select among these candidates, transforming originally linear trajectories into structured multi-step decision-making processes.

\section{Scaling Reinforcement Learning}
\label{sec:post_training}

Post-training with reinforcement learning has become a primary approach to elicit stronger reasoning capabilities.
Following LongCat-Flash-Thinking~\citep{longcat-thinking}, we adopt a unified multi-domain post-training pipeline, in which domain-specialized expert models are first trained under a shared framework and then consolidated into a single general model through both model-level and data-level merging.

Reinforcement learning with agentic reasoning requires:  (i) carefully prepared training setups, including scalable environment construction, high-quality cold-start data, and well-calibrated RL task sets; (ii) a dedicated infrastructure that can sustain high-throughput, asynchronous, and long-tailed multi-turn rollouts; and (iii) specialized training strategies that remain stable and effective across heterogeneous domains and varying difficulty levels.
In this section, we present a systematic pipeline that addresses these challenges and enables scalable agentic reinforcement learning.

\subsection{RL Preparation}
Before conducting large-scale reinforcement learning, several essential preparation steps are required.
Standard reinforcement learning relies on two core components: (i) a well-initialized policy that exhibits basic task-relevant behaviors, and (ii) a principled training task set that enables stable and effective learning.
In addition to these shared components, agentic reinforcement learning further requires a reliable and scalable environment foundation to support long-horizon, interaction-driven training.

\subsubsection{Environment}
\label{sec:post_training_env}

Traditional reasoning is largely environment-free, operating purely within a internally linguistic space.
In contrast, environments constitute a defining component of agentic systems, as they directly determine what an agent can perceive, how it can interact with external systems, and what actions are executable in the world.
We specifically introduce two particularly challenging types of environments in agentic tasks: agentic coding and agentic tool-use.
For agentic coding tasks, the environment takes the form of an executable code sandbox, which must simultaneously support complex real-world toolchains, ensure strict execution correctness, and remain stable and reproducible under large-scale parallel usage.
For general agentic tool-use tasks, the environment must model complex and diverse interactions across heterogeneous tools and databases in a consistent manner.
These requirements call for dedicated and domain-specific environment designs, which we describe in detail below.

\paragraph{Code Sandbox}
For each coding task, the agent must operate within an executable code sandbox and interact with various terminal tools in real time.
This setting introduces two major challenges: the need for efficient large-scale resource scheduling and the heterogeneity of tools and execution environments.
In large-scale training, the sandbox system must simultaneously ensure flexibility and robustness.
To address these challenges, we design a scalable execution sandbox system that provides unified tool interfaces with high-throughput interaction.
To handle the heterogeneity of real-world code execution settings, we consolidate commonly used tools—including search, file reading and writing, code editing, and shell execution—into a standardized environment interface.
To further support large-scale training, we implement a high-concurrency sandbox scheduler that asynchronously provisions and recycles sandbox instances, distributes tasks across workers, and executes thousands of sandboxes in parallel, thereby removing environment startup and blocking overhead from the critical training path.

\begin{figure}[t]
    \centering
    \includegraphics[width=0.95\textwidth]{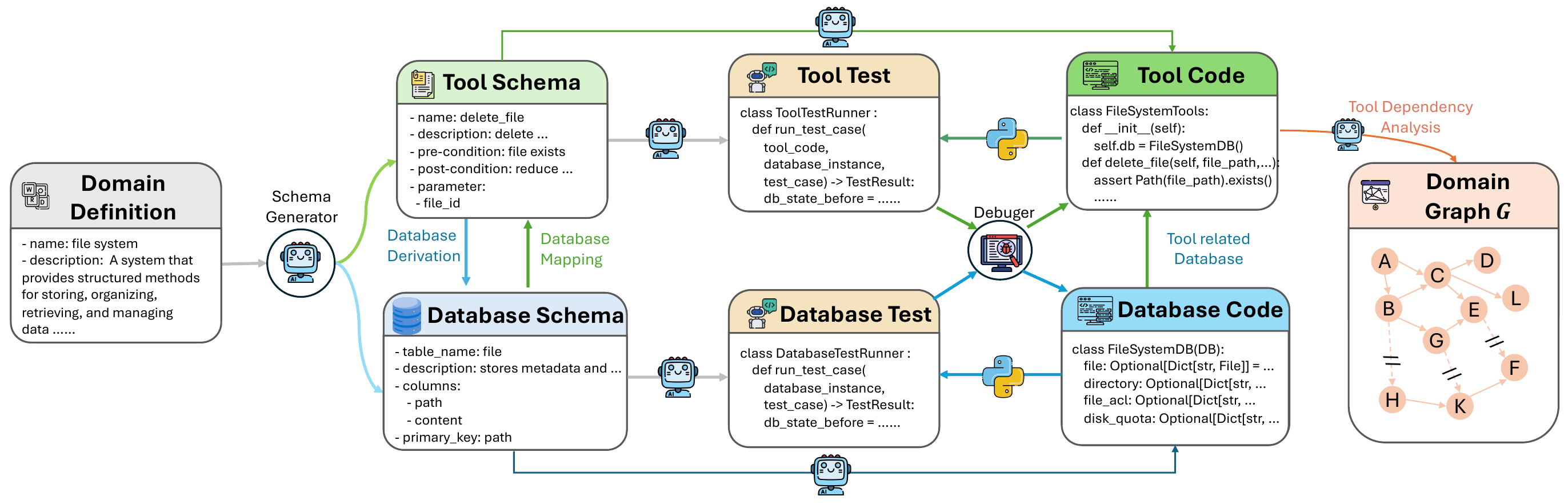}
    \caption{Automated construction of executable domain graph.
Starting from a high-level domain specification, the pipeline synthesizes domain-specific tools, generates corresponding databases schema and tool schema implementations. Building on these we construct a verified tool dependency graph that serves as the foundation for environment scaling.}
\label{fig:domain_env}
\end{figure}

\begin{figure}[t]
    \centering
    \includegraphics[width=0.95\textwidth]{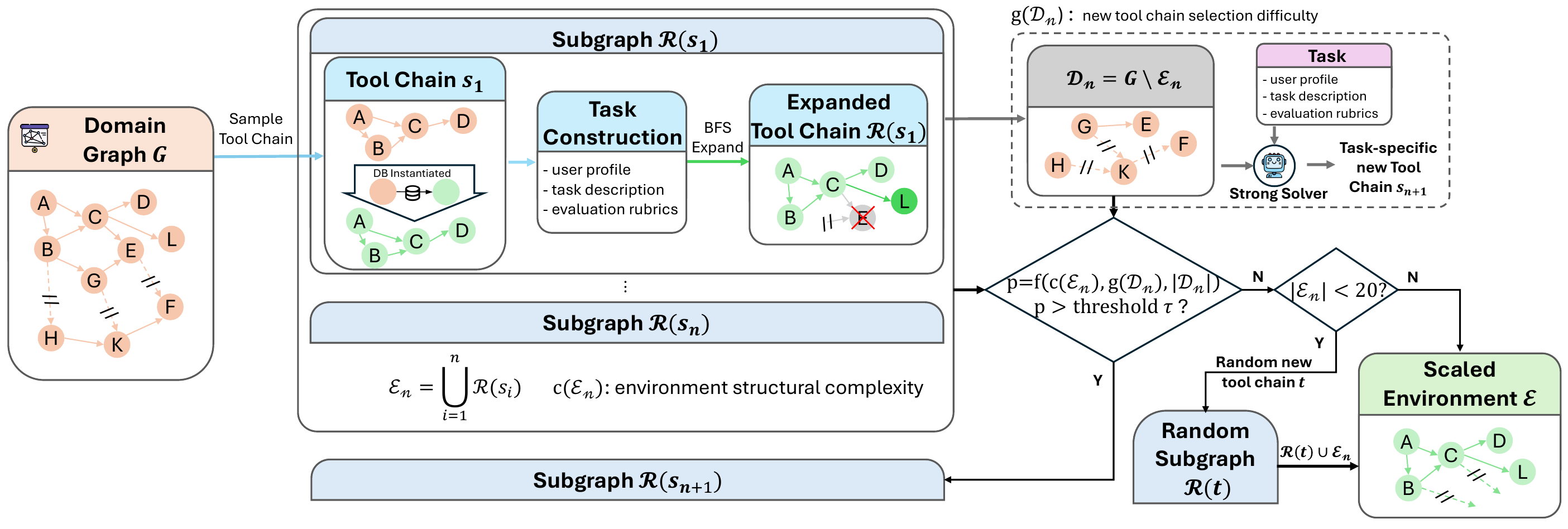}
    \caption{Verifiability-preserving environment expansion.
Environment is progressively expanded from a seed executable tool chain through controlled graph growth.}
\label{fig:env_scaling}
\end{figure}

\paragraph{Agentic Tool-Use Environment}
The core characteristics of advanced agentic reasoning capability is generalization, which is the ability to acquire effective behaviors in known environments and reliably transfer them to previously unseen settings.
To train such generalized agents, we posit that transferable agentic reasoning capability should emerge from exposure to diverse tool sets and interaction patterns.
While it is infeasible to design a single meta-environment that fully captures the complexity of real-world tool use, we find that a sufficiently diverse set of environments can provide a basis, enabling transferable and generalizable tool-use behaviors.
In practice, however, constructing such environments presents non-trivial challenges, as different domains impose distinct requirements on interaction interfaces, execution mechanisms, and evaluation criteria.

We design a fully automated pipeline that converts a high-level domain specification into an executable graph. As illustrated in Figure~\ref{fig:domain_env}, the pipeline synthesizes a domain-specific tool set from the domain definition and abstracts their functionalities into unified database schemas with explicit tool--database mappings. Based on this schema-level specification, it then automatically generates the database implementation and the corresponding tool code. To ensure correctness and robustness, the generated code is validated through unit testing and an auxiliary debugging agent. In practice, this pipeline achieves a success rate exceeding 95\% in transforming schema-level designs into fully executable tool implementations. Finally, we construct the tool dependency graph from the verified tool set, which serves as the basis for subsequent environment construction and expansion.

Using our automated pipeline, we construct a collection of domain-specific tool graphs $\mathcal{G}$ covering over 20 domains. Each tool graph $G \in \mathcal{G}$ contains more than 60 tools organized in a dense dependency graph, paired with a corresponding executable tool-specific database schema implementations, providing sufficient structural complexity for large-scale exploration and diverse environment construction.

We next introduce our pipeline for constructing executable environments based on domain graphs.
Given a domain-specific tool graph $G$, we first sample a moderate-size tool chain $s_1 \subset G$.
The sampling probability of previously selected tools is progressively reduced to promote task diversity.
For each tool in $s_1$, we sequentially instantiate the corresponding database states, ensuring that all required tool dependencies are fully satisfied.
After the tool-associated databases are constructed, we automatically generate a task grounded on this tool chain. 
The sampled $s_1$ serves as an executable tool chain for the generated task.
To mitigate potential human prior bias, the task generator was restricted to utilizing the full tool chain exclusively, without any deliberate task design.
Specifically, each generated task consists of three components: (i) task description, (ii) user profile, and (iii) evaluation rubrics.
To ensure reliable supervision, the evaluation rubrics are validated through multiple rounds of consistency checking, guaranteeing that any executable tool chain can be accepted as a correct solution and that incorrect or incomplete trajectories are consistently rejected.

We note that task difficulty arises from two complementary sources: interaction complexity and environment complexity.
Interaction complexity is ensured by synthesizing diverse user prompts that require varying degrees of clarification, planning, and multi-step interaction, encouraging the model to engage in rich and adaptive agentic behaviors.
Environment complexity is quantified using structural properties of the underlying tool graph, including the number of involved tool nodes and the connectivity density among them.

A natural way to increase environment complexity is to expand the initial tool chain $s_1$ into a larger subgraph $\mathcal{R}(s_1)$ by introducing additional tool nodes from $G$, together with maintaining their database instances.
Notably, after such expansion, the resulting environment may additionally contain multiple valid executable tool chains.
However, as the subgraph grows, maintaining cross-tool database consistency becomes increasingly challenging.
In particular, due to the constraints imposed by the tool dependency graph, the uncontrolled introduction of additional tool nodes may trigger a cascade of previously unmet dependencies, requiring extensive database augmentation and the maintenance of a large number of auxiliary tool nodes.
Once the database state becomes inconsistent under such expansion, tool invocations may yield unpredictable outcomes, causing even valid tool trajectories to fail execution.
This can lead to correct solutions being incorrectly judged as failures, thereby introducing biased negative rewards during training.
To mitigate this issue, instead of blindly injecting random tools, we expand environments by extending executable tool chains.
Specifically, starting from $s_1$, we perform a BFS-style expansion on the tool graph and only add a new tool node if all of its dependencies are already satisfied by previously instantiated tools, whose database states have been fully constructed.
This controlled expansion strategy allows us to progressively increase environment complexity while preserving executability and reliable supervision.

Through this controlled graph expansion, we extend $s_1$ into a more complex subgraph $\mathcal{R}(s_1)$. We denote the resulting extended environment as $\mathcal{E}_1 = \mathcal{R}(s_1)$.
Let $\mathcal{D}_1 = G \setminus \mathcal{E}_1$ denote the remaining unused tool nodes for $\mathcal{E}_1$.
A further design choice is whether to initialize another seed executable tool chain $s_2 \subseteq \mathcal{D}_1$ and extend it into $\mathcal{R}(s_2)$ to construct the environment:
\[
\mathcal{E}_{2} = \mathcal{R}(s_1) \bigcup \mathcal{R}(s_2),
\]
which to further increase environment complexity.
We make this decision based on (i) the structural complexity of the current environment, $c(\mathcal{E}_{1})$, (ii) the difficulty of identifying a new valid tool chain from the remaining graph, $g(\mathcal{D}_1)$, and (iii) the number of remaining unused nodes $|\mathcal{D}_1|$.
Concretely, given the current environment:
\[
\mathcal{E}_{n} = \bigcup_{i=1}^{n} \mathcal{R}(s_i),
\]
we define the remaining unused tool set as $\mathcal{D}_n = G \setminus \mathcal{E}_n$,
and sample a new seed chain $s_{n+1}$ from $\mathcal{D}_n$ with probability:
\[
p = f\big(c(\mathcal{E}_n), g(\mathcal{D}_n), \lvert \mathcal{D}_n \rvert\big),
\]
where $g(\mathcal{D}_n)$ is measured by the number of attempts required by a strong solver to discover an alternative valid tool chain from $\mathcal{D}_n$, and $f(\cdot)$ is a monotonic decision function controlling the environment growth process. 
In practice, when $p > \tau$, where $\tau$ is a predefined threshold, we initialize a new seed golden tool chain $s_{n+1}$ from $\mathcal{D}_n$ and extend it into $\mathcal{R}(s_{n+1})$.
To ensure a minimum level of environment complexity, we further introduce a fallback mechanism.
If the resulting expansion $\mathcal{E}_n$ contains only a small number of tool nodes, we randomly sample an additional moderate-size tool chain from $G$ and incorporate it into the environment, while ensuring the database consistency with current tool chains.
This guarantees that each constructed environment contains at least 20 tools, providing sufficient structural complexity for meaningful agentic interaction and exploration.
This automatic environment construction strategy allows us to progressively scale environment complexity while maintaining reliable supervision signals, enabling stable and effective large-scale agentic reinforcement learning.

\subsubsection{Initial Policy}
\label{sec:post_training_cold_start}

The cold-start stage serves a critical role in initializing the policy for subsequent reinforcement learning stage.
Instead of prioritizing immediate gains on standard benchmarks, our primary goal is to prime the model for effective large-scale exploration.
A high-quality cold-start policy must exhibit diverse reasoning patterns and stable interaction formats, ensuring that the subsequent RL stage can be conducted efficiently while retaining general thinking capability.
Consequently, we evaluate the cold-start model from a more fundamental perspective, focusing on: (1) its proficiency on the specific tasks designated for the RL stage, and (2) the diversity of its reasoning paths, as assessed through qualitative human inspection.
While we report pass@k performance on standard benchmarks to gauge underlying reasoning competence, it serves as a reference rather than the primary optimization target.

A major challenge in cold-start training lies in constructing high-quality data that can effectively prime the model for agentic behaviors.
For some domains, such as mathematics, general coding, and agentic code tasks, large-scale data sources are available in real-world.
In these cases, we curate and assemble trajectories from existing sources, and the core challenge lies in strict quality control and executability verification.
However, for most of the agentic tasks, such as search and tool-use, high-quality real-world trajectories are largely unavailable.
As a result, we rely on carefully designed data synthesis pipelines to construct cold-start trajectories.
We next describe how we instantiate these two complementary strategies for different agentic capabilities in detail.

\paragraph{General Thinking}
Agentic capability require strong thinking capability as a foundation.
We therefore design a strict data filtering pipeline to construct high-quality general thinking data.
Specifically, we adopt a K-Center-Greedy (KCG) selection algorithm guided by perplexity (PPL)~\citep{KCG1, KCG2, KCG3}.
Existing PPL-based filtering methods either ignore sequence length or average PPL over the full sequence~\citep{PPL1, PPL2}, which masks localized hard tokens and discards informative samples.
To address this, we introduce sliding-window PPL, which computes the maximum average PPL over all 512-token windows in a sequence, capturing peak model uncertainty without dilution from global averaging.
During KCG selection, we weight the distance between a candidate sample and the current selected set by its sliding-window PPL score.
This design jointly captures two complementary objectives: KCG preserves coverage over the original data distribution, while sliding-window PPL emphasizes samples that expose gaps in the model’s current reasoning capability.
Using this pipeline, we downsample 210K general thinking samples from a large-scale corpus.
Models trained on this subset outperform those trained on the full corpus on multiple reasoning benchmarks.

\paragraph{Agentic Coding}
For agentic coding tasks, large-scale software development platforms provide a rich source of real-world trajectories.
However, raw trajectories collected from such sources are often noisy, partially incorrect, or irreproducible, and ensuring their reliability and executability becomes a critical challenge.
To address this, we construct a strict curation pipeline for code-interaction trajectories. 
We require that all retained trajectories are fully executable and verifiable within reproducible environments, and we keep only those that correctly resolve the target issue while preserving existing functionality. 
To further avoid learning spurious behaviors, we apply fine-grained action-level filtering, removing erroneous, redundant, or speculative operations that do not contribute to correct problem solving. 
In addition, to preserve long-horizon reasoning patterns commonly observed in real debugging processes, we retain trajectories involving long and iterative debugging by compressing earlier steps, allowing long-horizon code reasoning to be maintained without length constraints.

\paragraph{Agentic Search}
For agentic search capability, constructing high-quality trajectories requires explicitly modeling multi-step evidence gathering and complete condition verification, while avoiding spurious shortcuts based on partial information. Such structured reasoning traces are rarely available in real-world search logs. 
We therefore construct synthetic reasoning trajectories that prioritize correctness, reasoning completeness, and robustness against shortcut behaviors.
We apply strict filtering to ensure trajectory quality, removing trivial cases and enforcing a consistent reasoning and tool-use format.
To avoid shortcut learning such as lucky guesses based on partial evidence, we require trajectories contains explicitly verification of all conditions specified in the query.
We further retain trajectories that involve long and iterative exploration by compressing earlier steps, allowing long-horizon reasoning to be preserved without length constraints.
Finally, we scale data collection and increase behavioral diversity by reusing rollout trajectories from subsequent reinforcement learning stages.

\paragraph{Agentic Tool-Use}
For agentic general tool-use capability, the main challenge lies in modeling complex, stateful interactions across heterogeneous tools and databases with diverse schemas and dependency structures. 
Such environments and interaction traces are also difficult to obtain or standardize from real-world sources.
We construct a scalable data synthesis pipeline built on top of our environment scaling pipeline that models realistic tool-use environments covering 33 representative domains by jointly defining domains, tool schemas, database states, and task objectives, and generating multi-step tasks grounded in structured tool dependencies.
To ensure diversity, we explicitly promote variability in three aspects: domain coverage, trajectory structure, and interaction length.
Each task admits multiple distinct correct tool-call trajectories, and interaction horizons range from short dialogues to long, multi-turn executions, capturing a wide spectrum of task difficulty and behavioral patterns.
To ensure data quality, all synthesized trajectories are strictly filtered using rubric-based outcome validation and turn-level quality control.
Only trajectories that correctly reach the target final state are retained. 
Within these trajectories, we apply turn-level loss masking to exclude low-quality turns—such as failed tool calls or format violations—from loss computation, ensuring the model learns only from correct actions while preserving the full interaction context.

\subsubsection{RL Task Set}
\label{sec:post_training_problem_set}

In reinforcement learning, the environment defines what the agent can do, while the task set determines what the agent is trained to do. Together, they shape how the agent explores the environment, allocates computation, and refines its policy through repeated rollouts.
A well-designed task set should be both informative and of appropriate complexity, so that it can provide effective learning signals without being either trivial or intractable.
A common practice is to first construct a diverse collection of environments paired with tasks instructions spanning multiple domains and difficulty levels, and then assess task suitability by evaluating the model’s pass rate. Based on this signal, tasks that are neither trivially solvable nor prohibitively hard are selected for reinforcement learning.
However, the availability of training tasks varies significantly across domains.
For some domains, such as coding, a large number of complex and high-quality tasks already exist and can be directly collected and curated.
In contrast, for domains like agentic search and tool-use, suitable tasks are scarce or do not exist in a readily usable form, making direct collection insufficient.
We therefore introduce principled synthesis pipelines to construct task sets spanning diverse complexity levels for agentic search and tool-use scenarios.

\paragraph{Agentic Search}
We identify two fundamental difficulty factors that characterize search problems: (i) multi-hop reasoning over relational entity chains, and (ii) reasoning over a single entity under multiple ambiguous constraints. Most complex search tasks can be viewed as compositions or iterative refinements of these two factors.
\begin{itemize}
    \item \textbf{Graph-based QA Synthesis}: 
    To model multi-hop reasoning difficulty, we construct graph-based question-answer tasks through a systematic pipeline that builds relational graphs from Wikipedia entities and generates challenging reasoning problems. Our approach begins by extracting low-frequency entities from Wikipedia as initial seed nodes, then iteratively expands the graph by sampling from the existing entity set, retrieving their Wikipedia pages, and incorporating related entities along with their corresponding relations until reaching a predefined size threshold. Once the graph is constructed, we sample multiple fixed-size connected subgraphs and use them to generate question-answer pairs. During question generation, we leverage large language models to create questions that correspond to the subgraph information, then deliberately obfuscate explicit details such as numerical values, entity names, geographical locations, and temporal markers to maximize reasoning complexity. To ensure quality and correctness throughout the pipeline, we employ an LLM-as-a-judge approach at critical steps including entity relation extraction, question generation, and obfuscation to maintain baseline accuracy. Finally, for each generated question-answer pair, we utilize an agent-based methodology to identify other potential correct answers and assess their validity, retaining only those pairs where the original answer is correct and all other identified potential answers are incorrect.
    \item \textbf{Agent-based QA Synthesis}: 
    To model ambiguity-driven difficulty, we present a scalable, efficient data synthesis pipeline, where multi-agent collaborative interactions are orchestrated by a Finite State Machine (FSM). Within this framework, an Entity Extraction Agent identifies representative long-tail entities and extracts their salient attributes, which serve as the foundational ground truth for synthesis question. Subsequently, a Question Synthesis Agent utilizes a random sampling of these attributes to formulate tailored questions. To ensure precision, a Verification Agent leverages search and browse tools to rigorously validate whether the ground truth satisfies all constraints specified in the question, thereby mitigating the risk of entity-question mismatches. For each validated question, the Answer Generation Agent utilizes search and browse tools to produce candidate answers. The Judgment Agent evaluates the alignment between these candidates and the predefined ground truth. When the Judgment Agent identifies a non-ground-truth answer that still satisfies the verification criteria, it signifies a multi-answer conflict. To resolve this, the system randomly incorporates additional attributes of the ground truth entity and triggers a re-synthesis of the question to ensure its uniqueness. This pipeline facilitates the high-throughput generation of diverse, high-quality QA pairs across multiple domains. Furthermore, it enables an automated difficulty grading mechanism based on the accuracy metrics of the Answer Generation Agent.
\end{itemize}

\paragraph{Agentic Tool-Use}
For agentic tool-use, our task set is directly constructed through the environment scaling pipeline described in Section~\ref{sec:post_training_env}. 
Each synthesized environment naturally defines a standalone task, and a collection of such environments constitutes the final task set.

\subsection{Scalable Asynchronous Agentic RL Framework}
\label{sec:post_training_rl_framework}

\begin{figure}[t]
    \centering
    \includegraphics[width=0.95\textwidth]{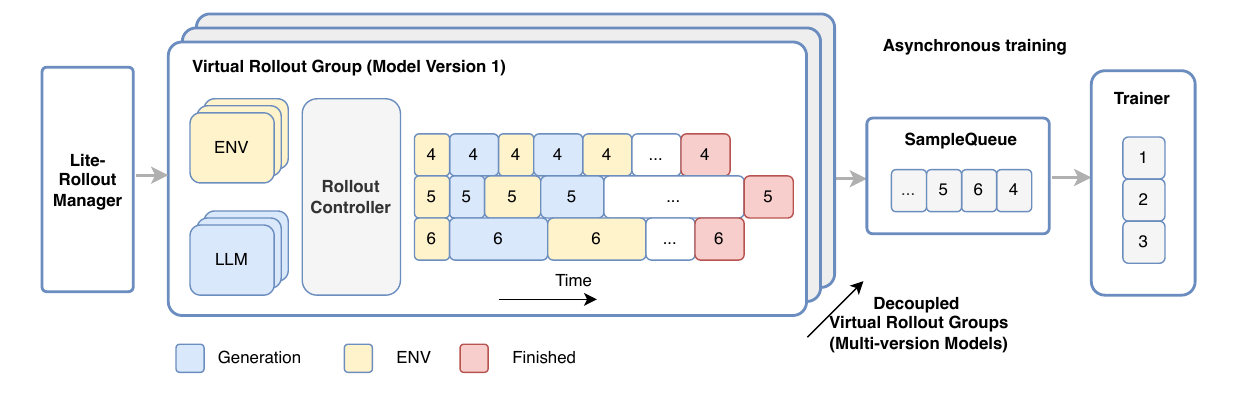}
    \caption{The Execution Workflow of Our Scalable Asynchronous Agentic RL Framework.}
\label{fig:scalable_async_framework}
\end{figure}

Compared with single-turn scenarios such as standard reasoning tasks, agentic training involves multi-turn interactions with variable environments (ENV) or tools, which pose new challenges to RL Infrastructure. During a typical multi-turn rollout stage, it repeatedly interleaves the LLM generation, environment execution, and reward assessment to construct the final trajectory for RL training. In this setting, trajectories are not only long-tailed skewed but also involve unpredictable and latency-skewed environment interactions, which lead to device underutilization under batch setting~\citep{fu2025areal,zhang2025agentrl,lu2025part}. Additionally, our production cluster consists of mid-range accelerators, especially with only around 60GB of available device memory. The hardware constraints, together with its software ecosystem, pose significant challenges to achieving stable and scalable agentic RL training. To address these problems, we extend our multi-version asynchronous training system, DORA (Dynamic ORchestration for Asynchronous rollout)~\citep{team2025introducing}, to fully support large-scale RL training in multi-turn agentic scenarios.

As illustrated in Figure \ref{fig:scalable_async_framework}, our controllers adopt a producer-consumer architecture consisting of a \texttt{RolloutManager} (which manages the rollout stage), a \texttt{SampleQueue} (which controls the sample staleness), and a \texttt{Trainer} (which manages the Experience-Maker and training stage). 
These components run on different nodes and are primarily responsible for logic control and coordination via Remote Procedure Call (RPC), while workers running on CPUs or accelerators execute the actual tasks. We propose several key techniques to enable efficient, scalable, and stable agentic RL training.

\paragraph{Fully Streaming Asynchronous Pipeline} 
To minimize device idleness in the agentic setting, we introduce a fully streaming asynchronous pipeline both within the rollout process and between rollout and training, based on streaming RPC~\citep{team2025introducing}. Within the rollout loop in the \texttt{RolloutManager}, we remove batch barriers, enabling LLM generation, environment execution, and reward computation to be executed on remote workers at the granularity of individual samples. This prevents accelerators from idling while waiting for batched ENV calls to complete. To further address the long-tailed generation problem with training stability, our DORA system enables multi-version asynchronous training, where trajectories generated by different model versions are immediately enqueued upon completion. Within a step, our multi-version generation instances continue rollout using multiple previous model versions, while the \texttt{Trainer} can initiate training as soon as its conditions are met, or elastically scale up additional generation instances for free extra throughput when training devices are idle.

\paragraph{Scaling to Large-scale Agentic Training} 
Our algorithm setting requires a large number of environments, e.g., up to 32,000 environments running across roughly 400 physical machines with thousands of accelerators. However, the intensive interactions between these environments lead to single-machine bottlenecks for our \texttt{RolloutManager} when scaling out, as each interaction typically involves a few CPU operations. To address this issue, we first decompose the original design into a \texttt{Lightweight-RolloutManager}, which manages global control metadata, and multiple \texttt{RolloutController}, each of which manages the lifecycle of a virtual rollout group in a data-parallel manner. A virtual rollout group consists of multiple trajectories and associated physical machines (including generation instances and ENV instances). Secondly, to flexibly schedule massive environments in our production jobs, we extend the PyTorch RPC framework~\citep{damania2023pytorch} to provide CPU-idleness-aware remote function invocation and object instantiation. This extension enables instantiating and executing remote environments on arbitrary or idle machines, which enables efficient deployment of massive environments.

\begin{figure}[t]
    \centering
    \includegraphics[width=0.8\textwidth]{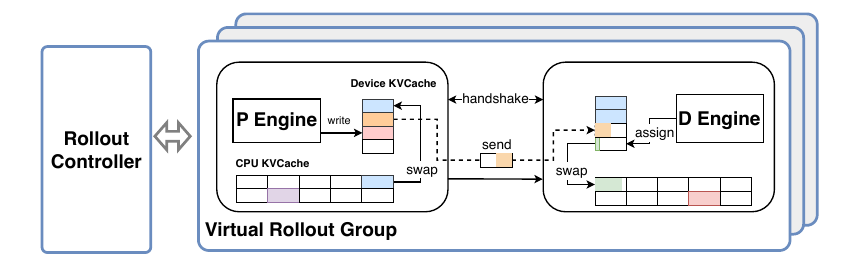}
    \caption{Workflow of Prefill-Decode Disaggregation with KV-cache Swapping. The rollout manager coordinates prefill (P) and decode (D) instances and collects final results from D instances. Upon receiving a new request, P and D perform an initial handshake and launch asynchronous chunked KV-cache transfers. When the device KV-cache usage reaches a given watermark, swapping is triggered and can be executed concurrently across both P and D instances.}
\label{fig:pd_disaggregation}
\end{figure}

\paragraph{PD Disaggregation with CPU Swapping} For efficient generation of \longcat, a 560B MoE model, on our accelerators, we employ a high degree of expert parallelism together with graph-level compilation for decode. However, in multi-turn agentic training, frequent incoming requests with long contexts lead to workload imbalance within an expert parallelism group: ranks assigned longer contexts consume disproportionately more computation and communication bandwidth, becoming the performance bottleneck. To address this issue, we introduce Prefill–Decode (PD) Disaggregation in RL training~\citep{zhong2024distserve,qin2024mooncake} as shown in Fig. ~\ref{fig:pd_disaggregation}. For generation instances, we deploy prefill nodes and decode nodes on separate device groups, allowing the decode execution graph to run without being interrupted by the prefill workloads of newly arriving requests. This prevents degradation in generation efficiency and maintains high throughput during multi-turn rollouts. PD Disaggregation, however, introduces additional challenges, including KV-cache transfer overhead and expensive recomputation overhead when the on-device KV-cache on the decode nodes is insufficient, especially on our accelerators. To mitigate KV-cache transfer cost, we aggregate KV-cache blocks in chunks level and enable asynchronous transmission between PD nodes. We enable overlapping between previous chunks with the computation of subsequent chunks to minimize the overhead of KV-cache transfer. To avoid recomputation caused by limited KV-cache memory on devices, we further introduce a CPU-resident KV-cache, which dynamically swaps KV-cache blocks in and out as needed. This design eliminates recomputation overhead due to insufficient on-device KV-cache capacity and helps sustain high throughput on our accelerators.

Overall, our agentic RL training framework achieves strong performance and stability at industrial scale on our 560B models, supporting tens of thousands of accelerators and environments. According to runtime statistics from our production jobs, our request load ratio\footnote{Request load ratio is a continuous value ranging from 0 to 1, aggregated over all generation devices across the entire rollout period. For example, our accelerator can accommodate approximately 4 requests when the maximum response length is set to 64K to avoid recomputation when considering only the available on-device KV-Cache blocks. The request load ratio will be 25\% if there is 1 request and 0\% when there are no requests on that accelerator, particularly in long-tailed generation scenarios.} is approximately 63\% throughout the entire rollout procedure. This metric quantifies the impact of long-tailed requests on generation throughput, with higher values indicating better utilization. We do not achieve full request load ratio during training because we limit new requests to the older model version to control average staleness at the expense of efficiency. Additionally, within a single step that may include multiple load-balancing operations, we employ a two-phase strategy: in the initial rollout phase before the first load-balancing when there are no long-tailed generation, we allow a higher number of requests per device (e.g., 8), and subsequently limit the requests per device to an optimal level (e.g., 4) to avoid recomputation and improve generation efficiency. In the future, we plan to adopt more optimistic staleness control strategies and explore staleness-aware stability techniques to enable more efficient asynchronous training. In conclusion, our multi-version asynchronous training system, DORA, is 2 to 4 times faster than synchronous training across our production jobs spanning different scenarios.

\subsection{RL Training Strategy}
\label{sec:post_training_rl_scaling}

Reinforcement learning has become a central mechanism for continuously improving a model’s reasoning capability.
Traditionally, RL training strategies mainly focus on stabilizing policy optimization, improving sample efficiency, and managing the exploration--exploitation trade-off under relatively homogeneous task distributions and single-step rollouts.
In these settings, tasks can exhibit widely varying difficulty levels, leading to highly imbalanced learning values across training samples.
Aside from curriculum learning, we dynamically allocate rollout budgets within each training step to make effective use of limited computational and training-time budgets, concentrating learning resources on high-value tasks.
To further improve training effectiveness, we additionally model verification as an auxiliary task that supports generation and accelerates optimization.

When extended to the agentic setting, reinforcement learning faces new challenges arising from multi-turn interactions and unpredictable environment feedback, which place stringent demands on the model’s effective context length.
To address this issue, we introduce a context management strategy that enables the model to support long-horizon trajectories under limited context windows, while preserving the most informative context.

When further extended to agentic tool-use scenarios, the training problem becomes substantially more challenging.
As mentioned earlier, generalization and robustness are particularly important in this setting.
To promote generalization, agentic tool-use environments are drawn from our environment scaling pipeline.
Tasks can span heterogeneous environments across diverse domains in a single batch, amplifying training instability and imbalance.
Under a careful co-design of training strategies and infrastructure support, we perform efficient and stable multi-domain environment training at scale.
Generalization to real-world settings remains challenging, as real-world environments are inherently imperfect.
To improve robustness, we explicitly incorporate environmental imperfections into the training process, enabling the model to learn resilient behaviors under non-ideal conditions.
Together, these designs form a unified training strategy that enables stable, efficient, and scalable agentic reinforcement learning.

\subsubsection{General Training Strategy}
In large-scale reinforcement learning, the training set spans tasks with highly diverse difficulty levels.
As a result, naive uniform treatment of all tasks often leads to inefficient learning.
To this end, we design a set of training strategies that are applied throughout all our reinforcement learning recipes.
Specifically, we introduce curriculum learning to progressively increase task difficulty, apply dynamic budget allocation to focus computation on the most informative tasks under the current model state.
We incorporate self-verification as an auxiliary task to further improve optimization efficiency and effectiveness.

\paragraph{Training Objective}
We adopt Group Sequence Policy Optimization (GSPO) as our training objective, as its empirical effectiveness on MoE models and provides more stable sequence-level optimization for long-horizon agentic trajectories.
Given an input $x$, we sample a group of $G$ trajectories $\{y_i\}_{i=1}^G$ from the old policy $\pi_{\theta_{\text{old}}}$ and optimize:
\begin{equation}
\mathcal{J}_{\text{GSPO}}(\theta)
=
\mathbb{E}_{x \sim \mathcal{D},\, \{y_i\}_{i=1}^G \sim \pi_{\theta_{\text{old}}}(\cdot|x)}
\left[
\frac{1}{G}
\sum_{i=1}^G
\min\!\left(
s_i(\theta)\,\hat{A}_i,\;
\mathrm{clip}\!\left(s_i(\theta), 1-\epsilon, 1+\epsilon\right)\hat{A}_i
\right)
\right],
\end{equation}
where $\epsilon$ is the clipping threshold.
Following~\cite{gspo}, we adopt group-based advantage estimation and define the importance ratio $s_i(\theta)$ at the sequence level based on normalized likelihoods.
We rely primarily on outcome-oriented supervision and relax penalties on long trajectories, allowing effective strategies to emerge naturally during training.

\paragraph{Curriculum Learning}
To improve learning effectiveness, we adopt a curriculum learning strategy that progressively structures the training process.
Concretely, our curriculum is organized along two complementary axes: task difficulty and capability requirement.
Task difficulty is quantified using the model’s pass rate estimated prior to optimization, where lower pass rates indicate more challenging tasks.
In parallel, we characterize tasks by the agentic capabilities they primarily exercise, such as basic tool invocation, multi-step planning, or autonomous decision making.
During early training stages, we prioritize tasks that are either easier to learn or expose capabilities that are expected to be autonomously reused by the agent when solving harder tasks.
As training progresses, we gradually shift toward tasks that are both more difficult and require more advanced combinations of agentic capabilities.
This two-dimensional curriculum allows the model to first acquire reusable agentic skills and then compose them to solve increasingly complex problems.
Empirically, this curriculum strategy improves the overall task pass rate and yields particularly significant gains on the most challenging tasks.
Our analysis in agentic settings reveals that these improvements stem from three main factors:
\begin{itemize}
    \item \textbf{Tool-use generalization}:  
    Skills acquired from simpler tasks—such as tool selection and constraint handling—transfer effectively to more complex scenarios, substantially reducing tool-call failures.

    \item \textbf{Interaction efficiency}:  
    Improved understanding of task instructions enables more thorough internal reasoning and reduces redundant clarifications or unnecessary tool invocations, thereby decreasing the number of interaction rounds.
    
    \item \textbf{Planning capability}:  
    Enhanced joint reasoning over multiple constraints such as time, location, and entities enables more direct task completion with fewer corrective iterations.
\end{itemize}

\paragraph{Dynamic Budget Allocation}

We further apply dynamic budget allocation within a training batch to prioritize tasks that provide higher learning value under the current model state.
We observe that tasks whose difficulty is well matched to the model’s current capability yield substantially higher learning gains under a fixed rollout budget.
Existing large-scale RL pipelines typically assign a uniform rollout budget to all tasks.
Recent research starts to explore adaptive rollout allocation~\citep{knapsack}, but most approaches rely on predefined value functions.
However, this assumption breaks down as the model’s capabilities evolve continuously during training, causing the set of tasks that provide the most informative learning signals changes accordingly.

To address this, we propose a dynamic rollout budget allocation strategy that adapts to the model’s real-time training state.
Specifically, we quantify the capability of the current policy $\pi_{\theta_t}$ by monitoring real-time training metrics $\mathbf{m}_t$ (e.g., pass rate). 
We map both the specific task $\tau_i$ and $\mathbf{m}_t$ via a dynamic value function:
\begin{equation}
    v_{i,t} = V(\tau_i \mid \pi_{\theta_t}, \mathbf{m}_t)
\end{equation}
where $v_{i,t}$ indicates the estimated value of task $\tau_i$, characterizing the model's evolving preference distribution over the task space.
Based on this value estimation, we employ a heap-based greedy algorithm to compute the rollout allocation that maximizes the aggregate learning value of the current training batch.
% As shown in Figure \ref{fig:dynamic}, dynamic budget allocation strategy consistently outperform uniform rollout.

\paragraph{Self-Verification}
Beyond using the model solely as an actor policy for generation, we additionally leverage the model as a verifier to assess the quality of its own on-policy trajectories.
We observe a notable asymmetry: even advanced reasoning models that are capable of generating high-quality trajectories often struggle to reliably assess the correctness of those trajectories in the absence of explicit ground-truth signals.
This motivates us to explicitly enhance the model’s self-verification capability and use it as an auxiliary signal to improve its reasoning ability in specific domains.

Concretely, we introduce on-policy self-verification as a dynamically activated training phase during reinforcement learning.
When the generator exhibits signs of stagnation or convergence to local optima, we trigger a verification stage in which the model evaluates its own rollout trajectories.
Compared to generation, verification is a comparatively easier task and typically yields higher rewards.
To further enhance the effectiveness of self-verification, we adopt a tailored training recipe.
Specifically, we ensure that verification is emphasized on challenging cases, and the influence of verification is coupled with the quality of the corresponding generated trajectories, so that the auxiliary signal encourages faithful improvement of generation rather than degenerate shortcut behaviors.
Empirically, introducing on-policy self-verification as an auxiliary task accelerates model convergence, leading to improved generation performance.

\subsubsection{Agentic Specific Strategy}
In agentic scenarios, the interaction pattern naturally shifts from single-turn generation to multi-turn trajectories that interleave model reasoning and tool invocations.
As task complexity increases, both the number of interaction turns and the length of tool responses grow, making the total context length highly variable and often difficult to control.
In practice, this frequently leads to context window overflow, truncated reasoning chains, and incomplete task execution.
Therefore, effective context management becomes a necessary component for reinforcement learning with tool use under limited context windows.
% The same context management strategy is also applied during evaluation, enabling reliable execution of long-horizon agentic trajectories beyond the nominal context length.

\paragraph{Context Management}
We design a hybrid context management strategy to support long-horizon trajectories under limited context windows.
Existing agentic models mainly adopt two strategies for context management:

\begin{itemize}
\item \textbf{Summary-based Management}:
When the cumulative context length exceeds a predefined token threshold, historical tool call results are distilled into a concise summary that replaces the original context for contextual continuity. Based on the framework of ReSum~\citep{wu2025resum}, we employ the model itself as the summary tool and conduct a series of comparative experiments with different token threshold values, ultimately identifying the optimal threshold of 80K tokens. More details can be found in Appendix \ref{sec:token_threshold_performance_evaluation}. Furthermore, to enhance the model’s summarization performance, we synthesize a high-quality dataset consisting of 15K samples specifically for the cold-start phase. Empirical results demonstrate that this yields approximately 3\% gains in accuracy.

\item \textbf{Discard-based Management}:  
When the context length exceeds a predefined threshold, the model will discard the entire or partial historical context, and then resume or restart the generation process based on the truncated context. Following DeepSeek-V3.2~\citep{DeepSeek-V3.2}, we adopt the \emph{discard-all} strategy in our work.

\end{itemize}

Combining the above two strategies, we design a hybrid context management method tailored for agentic reasoning.Specifically, we first apply summary-based compression whenever the context window exceeds our predefined limit of 80K tokens.When the interaction exceeds the maximum number of turns,  we trigger a \emph{discard-all} reset and restart generation with an initialized system and user prompt derived from the original question. 

\begin{figure}[t]
    \centering
    \includegraphics[width=0.6\textwidth]{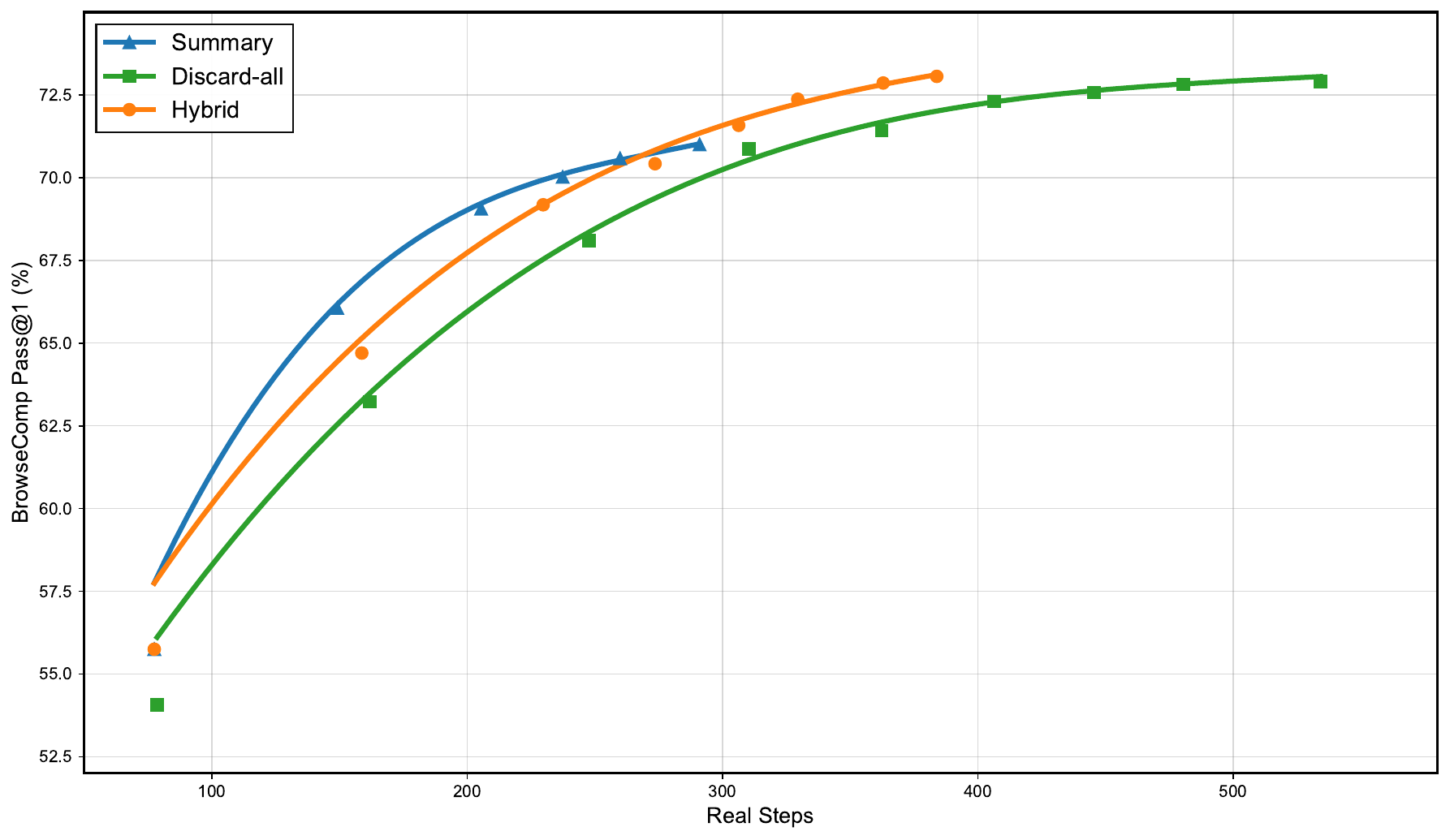}
    \caption{Pass@1 accuracy of BrowseComp with different context management strategies.}
    \label{fig:context_management_figure}
\end{figure}

We evaluate these three strategies on the BrowseComp benchmark~\citep{BrowseComp}. As illustrated in Figure~\ref{fig:context_management_figure}, under heterogeneous computational budgets, context management yields substantial performance improvements by enabling the model to scale test-time computation resources. Notably, the hybrid strategy outperforms the other two in most cases and demonstrates the highest efficiency throughout the experiment, starting from 55.8\% and reaching a peak of 73.1\%. This superior performance stems from dynamic switching between compression and reset, governed by context window and interaction turn constraints, which achieves a favorable trade-off between critical reasoning context retention and computational overhead control. Besides, we adopt a progressive discard schedule that gradually increases the discard threshold, allowing a gradual increase in the number of reasoning steps for more challenging samples.

\subsubsection{Training Strategy with Scaled Environment}
Introducing scaled environments further elevates the complexity of reinforcement learning beyond tool-augmented settings.
In this stage, we conduct large-scale training over environments that span multiple domains and exhibit high heterogeneity.
We expect that training over such diverse environments encourages the model to acquire transferable agentic behaviors that generalize beyond domain-specific patterns.
This heterogeneity introduces new challenges at both the algorithmic and system levels: the training process must simultaneously maintain cross-domain generalization, ensure stable optimization under highly diverse task distributions, and remain efficient under long-tailed and imbalanced environment workloads.
To address these challenges, we adopt a multi-domain environment training paradigm that jointly optimizes across diverse environments within each training batch.
To ensure both stability and scalability, we need co-design of training strategies and our asynchronous infrastructure.
Inspired by the environment scaling construction procedure, we notice that real-world environments are inherently imperfect, 
Accordingly, we explicitly introduce environmental noise during training to improve robustness to heterogeneous and unreliable environment feedback.

\paragraph{Multi-Domain Environment Training}
\begin{figure}[t]
    \centering
    \includegraphics[width=0.8\textwidth]{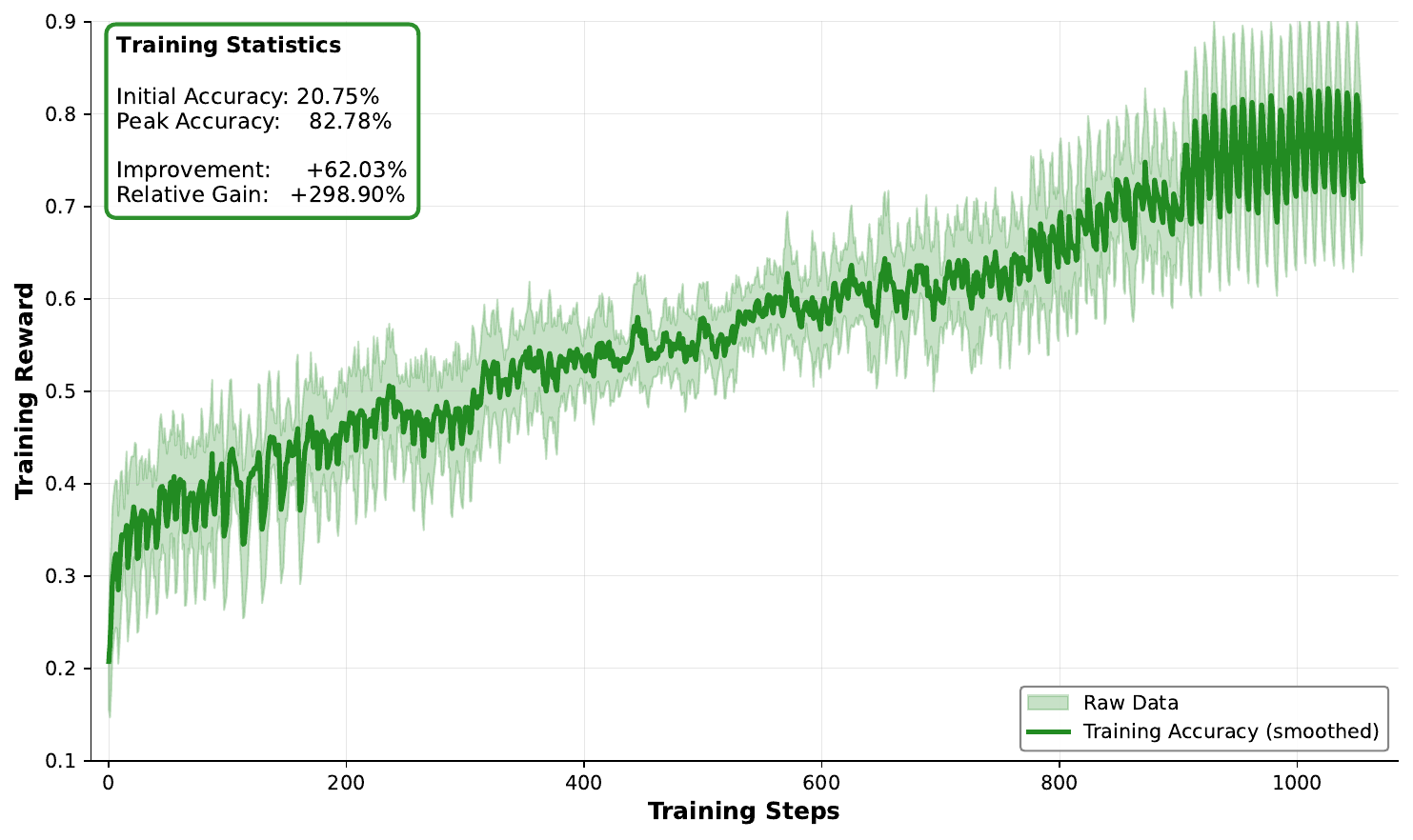}
    \caption{Training reward of \longcat during large-scale multi-environment agentic RL.}
    \label{fig:agentic_training_reward}
\end{figure}

\begin{figure}[t]
    \centering
    \includegraphics[width=0.9\textwidth]{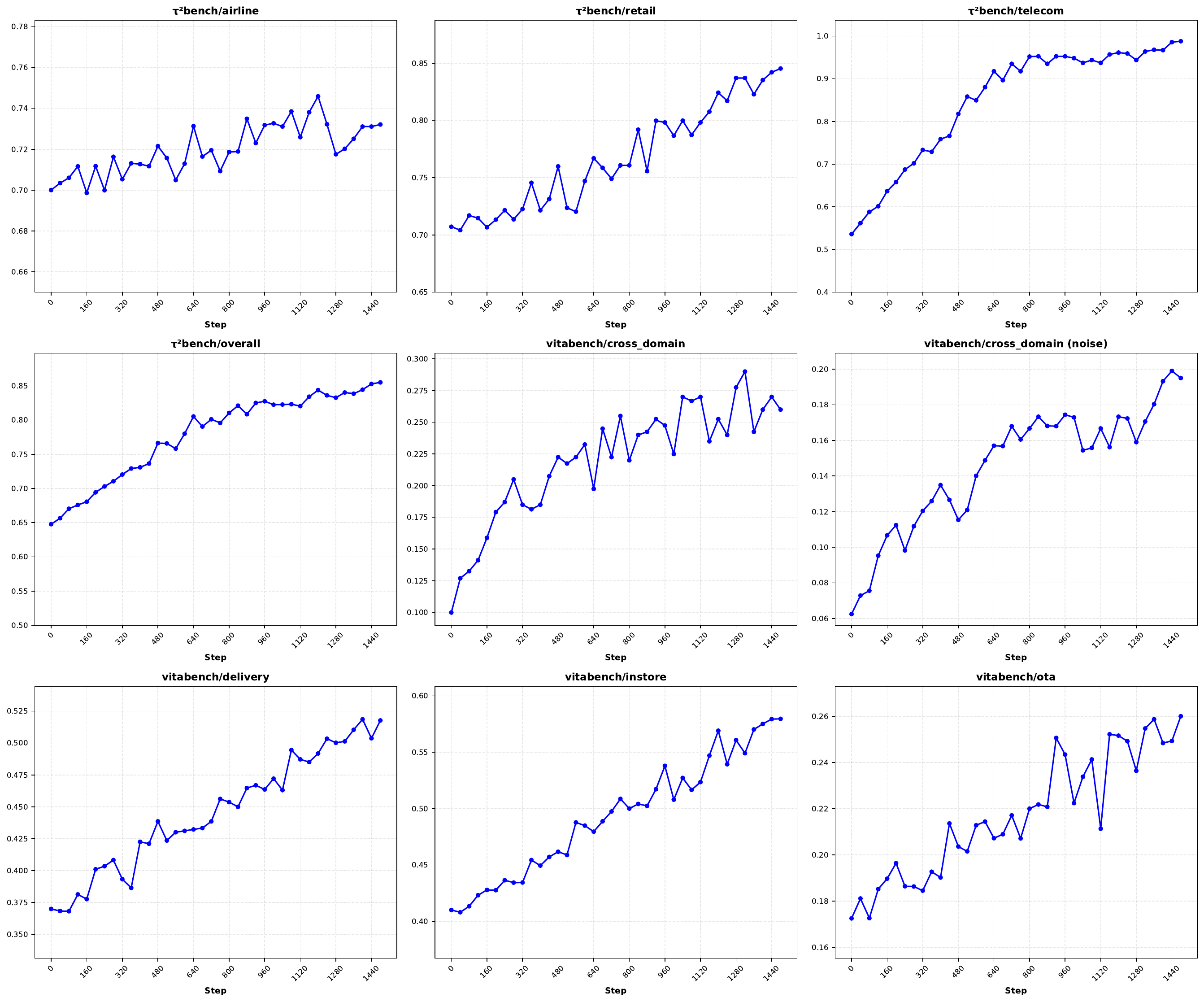}
    \caption{Agentic benchmark performance of \longcat during RL training using exclusively synthetic general agentic data.}
    \label{fig:agentic_benchmark_curves}
\end{figure}

To improve cross-environment generalization and training stability, we adopt a multi-domain environment training strategy that jointly optimizes across diverse environments within each training batch.
Through environment scaling, we construct tens of thousands of environments spanning more than 20 domains, providing broad coverage of heterogeneous interaction patterns.

Training under this setting introduces non-trivial system challenges. 
Specifically, to maintain training stability, it is necessary to ensure that all domains contribute comparably to the overall training process, while preventing any single training batch from being dominated by a small subset of domains.
This algorithm constraint significantly degrades the efficiency of our DORA system, as it breaks the design principle of asynchrony: the trainer may be forced to wait for slow or rare long-tailed domains to produce enough samples, while faster domains accumulate excessive rollout trajectories, leading to scheduling bubbles and device underutilization.
To mitigate this issue, we support configuring separate oversampling ratios for different data types and domains. In practice, we increase the rollout quota for more challenging or low-throughput domains, allowing them to contribute sufficient samples without blocking the overall pipeline, while faster domains are sampled at a lower effective rate. This design relaxes the strict per-batch balancing constraint and maintains the high-throughput property of asynchronous training using DORA, while still maintaining a roughly balanced data mixture at the training stage with training convergence guarantee.

To ensure the compatibility of dynamic budget allocation in this training setting with our asynchronous infrastructure, we introduce an oversampling coefficient for each task based on its historical pass rate. 
Tasks with lower success rates are assigned higher oversampling coefficients, effectively allocating more rollout budget to more challenging tasks. 
Concretely, each task is duplicated into multiple groups according to its oversampling coefficient, and each group independently computes advantages during training. This design approximates dynamic budget allocation while preserving the simple and fully asynchronous scheduling behavior of DORA, achieving dynamic token budget control with minimal system complexity.

We present the training reward curves of \longcat in Figure~\ref{fig:agentic_training_reward} and the corresponding performance on agentic benchmarks at different training steps in Figure~\ref{fig:agentic_benchmark_curves}. The training reward in Figure~\ref{fig:agentic_training_reward}  exhibits a stable and consistent increasing trend throughout training, indicating that our algorithm–infrastructure co-design effectively ensures training stability at scale. 
Moreover, the performance of agentic benchmarks in Figure~\ref{fig:agentic_benchmark_curves} demonstrates strong generalization across multiple benchmarks, which validates the effectiveness of our environment synthesis pipeline.
Empirically, multi-domain environment training achieves a higher average task completion rate across environments, with especially significant improvements on the most challenging environments.
Moreover, the model attains strong performance in randomly generated environments, demonstrating strong generalization.

\paragraph{Robust RL}
\begin{table}[t]
\centering
\caption{Performance comparison across different training strategies.}
\label{tab:robust_rl_results}
\scalebox{0.8}{
\begin{tabular}{lccc}
\toprule
Dataset & ColdStart & Training w/o Noise & Training w/ Noise \\
\midrule
VitaBench (Avg@4)           & 10.0 & 28.6 & \textbf{29.3} \\
VitaBench-Noise (Avg@4)     & 6.3  & 13.3 & \textbf{20.5} \\
\midrule
Tau2Bench (Avg@4)           & 78.8  & 87.1 & \textbf{88.2} \\
Tau2Bench-Noise (Avg@4)     & 58.8  & 62.2 & \textbf{67.1} \\
\bottomrule
\end{tabular}
}
\end{table}

We explicitly incorporate environmental imperfections into the training process to improve robustness. 
Existing agentic models suffer from significant performance degradation when deployed in previously unseen or imperfect environments. 
This issue largely stems from a common assumption in current agentic training paradigms: agents are typically trained with carefully curated instructions and interact with stable, well-controlled environments.
In contrast, real-world environments are inherently imperfect. Users exhibit diverse interaction styles and unpredictable behaviors, while tools may fail, return noisy outputs, or produce incomplete results due to various external factors.
Rather than assuming idealized environments during training and relying on agents to adapt post hoc, we systematically analyze real-world noise and design a automated pipeline to explicitly incorporate environmental imperfections into the training process.
To avoid introducing unreliable or misleading reward signals, we ensure that the injected imperfections do not invalidate task solvability, but instead increase the difficulty and stochasticity of the interaction process. 
Concretely, we model two major sources of interaction noise in real-world agentic scenarios: instruction noise, which captures ambiguity and variability in user interaction patterns, and tool noise, which simulates execution failures, inconsistent responses, and partial results from external tools.

During agentic reinforcement learning, we introduce these noises progressively using a curriculum-based strategy.
We measure the robustness of an agentic model as the performance gap between perfect and imperfect environments on the same task.
Starting from mild perturbations, we gradually increase noise difficulty and diversity as the model demonstrates sufficient robustness at the current level.
This adaptive process ensures that training remains informative rather than overwhelming, and avoids inefficient exploration of excessively noisy regimes.

Table~\ref{tab:robust_rl_results} presents an ablation study of our robust training strategy under imperfect environments. 
The results show that training with noise achieves comparable or even slightly better performance on standard agentic benchmarks, while yielding substantial improvements under noisy and imperfect conditions.

\section{Test-Time Scaling Through Heavy Thinking}
\label{sec:post_training_test_time}

\begin{figure}[t]
    \centering
    \includegraphics[width=0.9\textwidth]{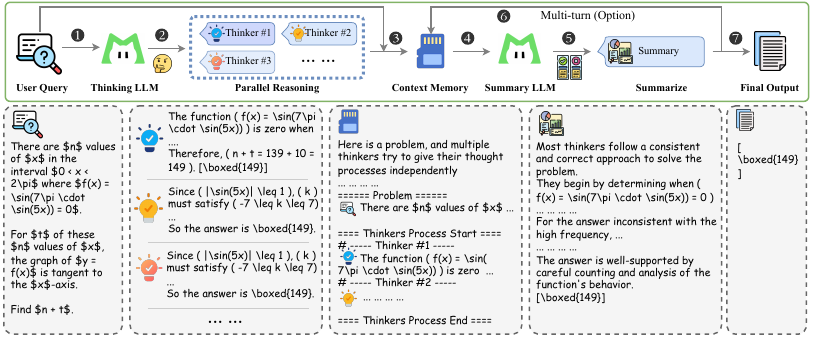}
    \caption{Framework of heavy thinking mode.}
\label{fig:heavy_mode}
\end{figure}

Test-time scaling (TTS) has emerged as an effective paradigm for improving model performance on complex reasoning tasks by expanding computation during inference.
Recent advances demonstrate that increasing reasoning depth—through long chains of thought with self-reflection allows models to iteratively refine their reasoning process~\citep{openai_o1, Deepseek-R1, gemini25pro}.
In parallel, approaches such as self-consistency and Monte Carlo Tree Search (MCTS) expand computation along the width dimension, exploring multiple reasoning trajectories to better approximate the model’s reasoning boundary.
More recently, several frontier models have introduced heavy thinking~\citep{Gemma3, gpt5, Kimi_K2_web_doc, pacore}, which aim to jointly scale both reasoning depth and width at test time.
Empirically, these modes outperform strategies that scale only depth or width.
However, the concrete implementation details of such heavy thinking remain largely undisclosed, limiting their reproducibility and systematic study.

\begin{figure}[t]
    \centering
    \includegraphics[width=0.9\textwidth]{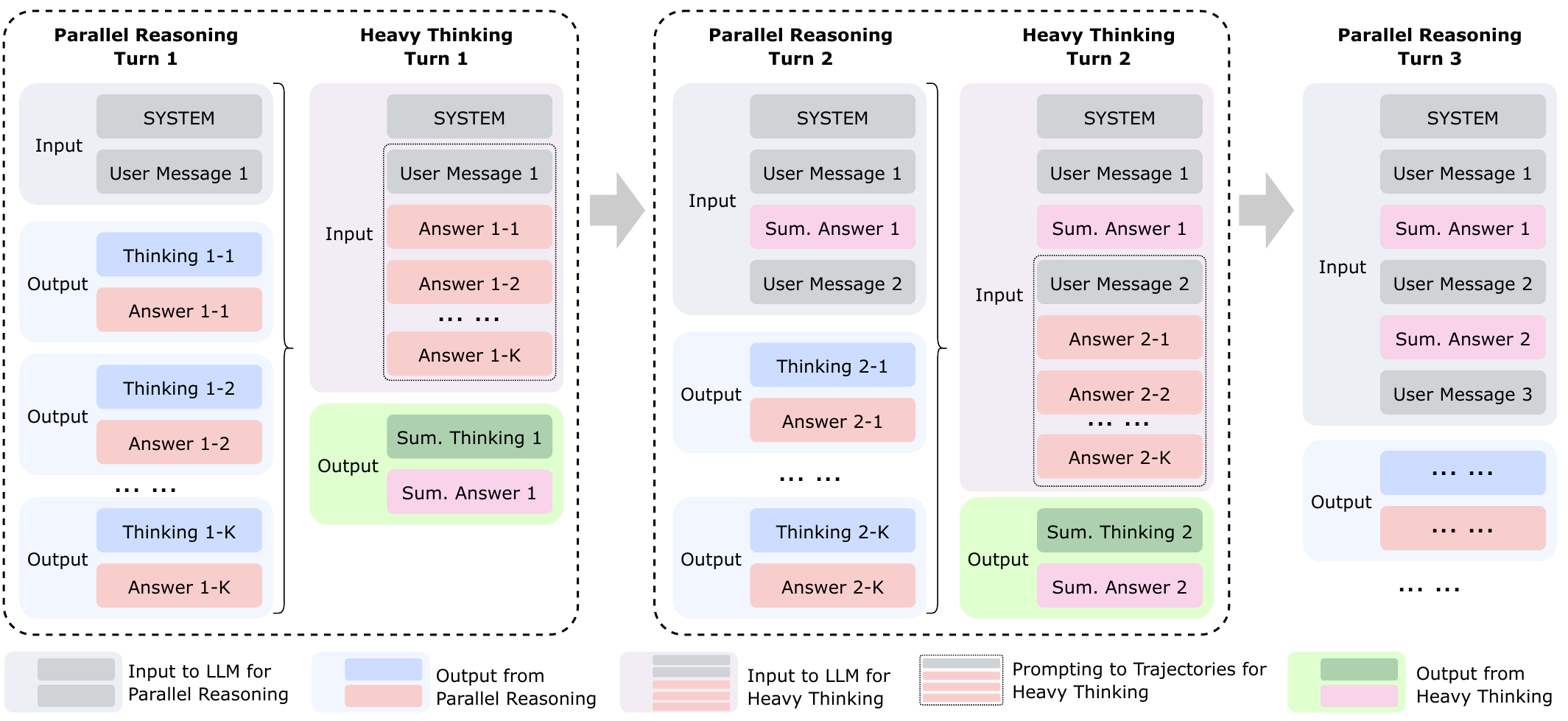}
    \caption{The context message management for parallel reasoning and heavy thinking.}
\label{fig:heavy_message}
\end{figure}

To further unlock reasoning capability and push beyond existing performance ceilings, we propose a simple and effective heavy thinking framework that decomposes test-time computation into two complementary stages: parallel reasoning and heavy thinking.
As shown in Figure \ref{fig:heavy_mode}, in the first stage, we allow a thinking model to perform generation in parallel, producing multiple candidate reasoning trajectories to expand the breadth of exploration.
In the second stage, we utilize a summary model to conduct reflective reasoning over these trajectories, synthesizing their intermediate reasoning and outcomes to arrive at a final decision.
To support tool use and multi-turn conversation scenarios, we also introduce a context memory module to store the history of messages. As shown in Figure~\ref{fig:heavy_message}, in each turn, the summary model will receive the history messages from the parallel reasoning stage to perceive context. 
We design a specific prompt template to organize the permutations of the parallel trajectories (only retain answer content) at the current turn, and elicit the summary model to generate the final response, which aims to aggregate or refine the answers derived from the parallel reasoning stage.
We also constrained the final output response of the summary model to maintain consistency with the style and format of the parallel reasoning stage, enabling direct concatenation of the response from the summary model with the message history.
Notably, the thinking and the summary module can either share the same model parameters or be instantiated as distinct models.

To further enhance performance, we also introduce an additional RL stage specifically tailored to the summary phase.
Empirically, we find that heavy thinking is effective across a wide range of settings, including long chain-of-thought reasoning, tool-integrated reasoning, and fully agentic tool-use scenarios.
By enabling test-time computation to scale adaptively along both reasoning depth and width, heavy thinking consistently outperforms self-consistency, with its performance advantage becoming increasingly pronounced as the test-time computational budget grows.

\section{Evaluation}
\label{sec:evaluation}

\begin{table*}[t]
\setlength{\tabcolsep}{4pt}
\centering
\caption{Performance (\%) comparison across multiple benchmarks (Best in \textbf{bold}, best of open-source in \underline{underlined}).$\dagger$ indicates the score is from external reports. $\ddagger$ indicates the score using our heavy mode. $^*$ indicates the w/ tools result is unavailable, and thus the corresponding w/o tools result is reported instead. Regarding BrowseComp, the performance is reported both without and with the context management technique.}
\label{tab:main_results}
\resizebox{0.98\textwidth}{!}{
\begin{tabular}{@{}l| c c c c c c c c@{}}
\toprule
\multirow{3}{*}{\textbf{Benchmark}} 
& \multicolumn{4}{c}{\textit{Open-Weights Reasoning Models}} 
& \multicolumn{3}{c}{\textit{Closed-Weights Reasoning Models}} 
& \textit{Ours} \\
\cmidrule(lr){2-5} \cmidrule(lr){6-8} \cmidrule(lr){9-9}
& DeepSeek-V3.2- & Kimi-K2- & Qwen3-235B-A22B- & GLM-4.7- 
& Claude-Opus-4.5- & Gemini-3-Pro & GPT-5.2- & \textcolor{mygreen}{LongCat-Flash-} \\
& Thinking & Thinking & Thinking-2507 & Thinking 
& Thinking &  & Thinking-xhigh & \textcolor{mygreen}{Thinking-2601} \\
\midrule
Architecture & MoE & MoE & MoE & MoE & - & - & - & MoE \\
\# Total Params & 671B & 1T & 235B & 355B & - & - & - & 560B \\
\# Activated Params & 37B & 32B & 22B & 32B & - & - & - & 27B \\

% ===================== Mathematical Reasoning w/ Tools =====================
\midrule
\rowcolor{mygray}
\multicolumn{9}{c}{\textit{Mathematical Reasoning w/ Tools}} \\
AIME-25$_\text{(Avg@16)}$ 
& 93.5$^*$ & 99.1$^\dagger$ & 92.6$^*$ & 95.3$^*$ & \textbf{100.0} & 99.8 & \textbf{100.0} & \underline{99.6} / \textbf{100.0}$\ddagger$ \\
HMMT-25$_\text{(Avg@16)}$ 
& 93.5$^*$ & 95.1$^\dagger$ & 83.9$^*$ & \underline{98.1}$^*$ & 98.6 & \textbf{99.8} & 99.6 & 93.4 / 97.5$\ddagger$ \\
IMO-AnswerBench$_\text{(Avg@4)}$ 
& 77.7$^*$ & 78.7$^*$ & 73.0$^*$ & 84.0$^*$ & 82.8 & 86.7 & - & 78.6 / \textbf{86.8}$\ddagger$ \\
AMO-Bench EN$_\text{(Avg@16)}$ 
& 51.9$^*$ & 56.0$^*$ & 47.8$^*$ & 62.4$^*$ & 66.0 & \textbf{72.5} & - & 61.6 / \underline{66.0}$\ddagger$ \\
AMO-Bench CH$_\text{(Avg@16)}$ 
& 52.0$^*$ & 51.8$^*$ & 28.8$^*$ & 35.1$^*$ & 67.7 & \textbf{74.9} & - & 56.8 / \underline{67.5}$\ddagger$ \\

% ===================== Agentic Search =====================
\midrule
\rowcolor{mygray}
\multicolumn{9}{c}{\textit{Agentic Search}} \\
BrowseComp$_\text{(Pass@1)}$
& 51.4$^\dagger$ / 67.6$^\dagger$ & - / 60.2$^\dagger$ & -  & 52.0$^\dagger$ / 67.5$^\dagger$ 
& -  & -  & \textbf{65.8}$^\dagger$ / - & \underline{56.6} / \textbf{73.1} \\
BrowseComp-zh$_\text{(Pass@1)}$ 
& 65.0$^\dagger$ / - & - / 62.3$^\dagger$ & -  & 66.6$^\dagger$ / -
& -  & -  & -  & \textbf{69.0} / \textbf{77.7} \\
RW Search$_\text{(Pass@1)}$      
& 74.0 & 63.0 & 20.5 & 69.0
& 75.5 & 74.5 & \textbf{82.0} / - & \underline{79.5} \\

% ===================== Agentic Tool Using =====================
\midrule
\rowcolor{mygray}
\multicolumn{9}{c}{\textit{Agentic Tool Using}} \\
$\tau^2$-Retail$_\text{(Avg@4)}$ & 81.8$^\dagger$ & - & 71.9$^\dagger$ & - & \textbf{88.9}$^\dagger$ & - & 82.0$^\dagger$ & \underline{88.6} \\
$\tau^2$-Airline$_\text{(Avg@4)}$ & 63.8$^\dagger$ & - & 58.6$^\dagger$ & - & - & - & - & \textbf{76.5} \\
$\tau^2$-Telecom$_\text{(Avg@4)}$ & 96.2$^\dagger$ & - & 47.3 & - & 98.2$^\dagger$ & - & 98.7$^\dagger$ & \textbf{99.3} \\
$\tau^2$-Avg$_\text{(Avg@4)}$ & 80.6 & 74.3$^\dagger$ & 59.3 & 87.4$^\dagger$ & 82.4 & \textbf{90.7}$^\dagger$ & 80.6 & \underline{88.2} \\
% $\tau^2$-Retail-Noise$_\text{(Avg@4)}$ & 42.8 & 48.5 & 45.4 & 46.7 & 43.6 & \textbf{52.2} & \underline{49.5} & 49.1 \\
% $\tau^2$-Airline-Noise$_\text{(Avg@4)}$ & \underline{64.1} & 59.5 & 47.0 & 63.0 & 61.5 & \textbf{65.0} & 59.0 & 62.5 \\
% $\tau^2$-Telecom-Noise$_\text{(Avg@4)}$ & 85.5 & 81.4 & 40.6 & \underline{88.2} & 73.0 & 54.8 & 86.6 & \textbf{89.7} \\
$\tau^2$-Noise$_\text{(Avg@4)}$ & 64.1 & 63.1 & 44.3 & 66.0 & 59.4 & 57.3 & 65.0 & \textbf{67.1} \\
VitaBench$_\text{(Avg@4)}$ & 24.0 & 12.8 & 14.5 & 18.3 & 28.5 & \textbf{31.5} & 24.3 & \underline{29.3} \\
VitaBench-Noise$_\text{(Avg@4)}$ & 14.0 & 9.2 & 6.5 & 10.8 & 20.3 & \textbf{20.8} & 19.0 & \underline{20.5} \\
% Toolathon & 32.4 & 16.6 & 9.2 & 18.5 & - & - & - & - \\
Random Complex Tasks$_\text{(Avg@4)}$ & 32.5 & 29.7 & 28.3 & 25.3 & 32.6 & 32.5 & 17.2 & \textbf{35.8} \\

% ===================== General =====================
\midrule
\rowcolor{mygray}
\multicolumn{9}{c}{\textit{General QA}} \\
HLE text-only (w/o tools) & 24.1 & 24.4 & 17.8 & \underline{26.9} & 32.0 & \textbf{40.3} & 34.5$^\dagger$ & 25.2 \\
GPQA-Diamond$_\text{(Avg@16)}$ & \underline{86.9} & 85.4 & 80.5 & 84.9 & 86.9 & 91.9 & \textbf{92.9} & 80.5 / 85.2$\ddagger$ \\

% ===================== Coding =====================
\midrule
\rowcolor{mygray}
\multicolumn{9}{c}{\textit{Coding}} \\
LCB (24.08-25.05)$_\text{(Avg@4)}$ & 82.4 & 75.1 & 76.2 & \underline{84.8} & 82.8 & \textbf{88.1} & - & 82.8 \\
OJBench$_\text{(Pass@1)}$ & 41.8 & 42.3 & 35.6 & \underline{44.6} & 46.7 & \textbf{61.2} & - & 42.2 \\
OIBench EN$_\text{(Pass@1)}$ & 43.3 & 39.0 & 36.8 & 30.8 & 50.0 & \textbf{58.2} & - & \underline{47.7} \\
SWE-bench Verified$_\text{(Avg@5)}$ & 73.1 & 71.3 & - & \underline{73.8} & \textbf{80.9} & 76.2 & 80.0 & 70.0 \\

% % ===================== Safety =====================
% \midrule
% \rowcolor{mygray}
% \multicolumn{9}{c}{\textit{Safety}} \\
% Harmful        & - & - & - & - & - & - & - & - \\
% Criminal       & - & - & - & - & - & - & - & - \\
% Misinformation & - & - & - & - & - & - & - & - \\
% Privacy        & - & - & - & - & - & - & - & - \\
\bottomrule
\end{tabular}
}
\end{table*}

\subsection{Benchmarks and Configurations}
Our evaluation covers five aspects of model capability: mathematical reasoning, agentic search, agentic tool use, general reasoning, and coding.

\paragraph{Mathematical Reasoning}
We evaluate mathematical reasoning using standard Olympiad-level benchmarks, including AIME 2025~\citep{aime25}, HMMT 2025 (February)~\citep{HMMT25}, and IMO-AnswerBench~\citep{IMO-AnswerBench}.
Aside from these standard benchmarks, we introduce AMO-Bench~\citep{AMO-Bench}, the most challenging dataset among existing Olympiad-level benchmarks.
It contains 50 problems designed by human experts with both English and Chinese versions, enabling analysis of cross-lingual mathematical reasoning.
We release the English version of AMO-Bench and the evaluation scripts\footnote{\url{https://github.com/meituan-longcat/AMO-Bench/}}.
Due to the limited size of these datasets, we report Avg@k metrics, using k=4 for IMO-AnswerBench and k=16 for all other math benchmarks.
We primarily focus on tool-integrated reasoning performance, where the tool refers to code execution.
For external models with code execution support, we enable this functionality during evaluation.
For models without code support, we report either official results (if available) or their performance without tools.

\paragraph{Agentic Search}
We evaluate agentic search capability on BrowseComp~\citep{BrowseComp} and BrowseComp-ZH~\citep{BrowseComp-zh}, and report results under both settings with and without context management.
For BrowseComp-zh, we notice some errors in the original annotations and manually revised the answers for 24 cases\footnote{\url{https://github.com/AGI-Eval-Official/BrowseComp-ZH-revised}}.
We attempted to reproduce the reported results of both open-source and closed-source models on these benchmarks under our agentic search framework.
However, we observed consistently lower performance than the reported numbers.
As a result, for external models, we use results from their official reports.
To enable a fair and controlled comparison of search capability, we additionally construct RWSearch\footnote{\url{https://github.com/AGI-Eval-Official/RW-Search}}, a challenging agentic search benchmark which consists of 200 real-world search queries that require complex reasoning and multi-step information retrieval.
All models are evaluated on RWSearch without context management, ensuring a fair comparison.

\paragraph{Agentic Tool-Use}
We evaluate agentic tool-use capability on $\tau^2$-Bench~\citep{tau2-bench}, VitaBench~\citep{vita-bench}, $\tau^2$-Noise, Vita-Noise, and Random Complex Tasks. 
For $\tau^2$-Bench, we observe that the default user simulator occasionally exhibits abnormal behaviors that introduce uncontrolled noise into the evaluation. 
To address this issue, we replace the original simulator with GPT-4.1 and adjust the prompting strategy accordingly~\citep{Claude-Opus-4.5, GPT-5.2, Gemini3}. 
For the airline subset of $\tau^2$-Bench, we further identify several annotation and environment issues that can lead to spurious failures and unreliable evaluation~\citep{Claude-Opus-4.5, GPT-5.2, Gemini3}. 
We therefore evaluate all models on a fixed and cleaned version of the airline subset, in which 19 problematic cases are corrected.
All these modifications have been publicly released for reproducibility\footnote{\url{https://github.com/AGI-Eval-Official/tau2-bench-revised}}.
For VitaBench, we update the evaluation setup by upgrading the verifier model to the strongest publicly available version and adopting a stricter evaluation criterion, thereby further improving the reliability of the benchmark.
We have publicly released this updated version of VitaBench\footnote{\url{https://github.com/meituan-longcat/vitabench }}.

We next introduce the construction and evaluation protocols for $\tau^2$-Noise, Vita-Noise, and Random Complex Tasks.
\begin{itemize}
    \item \textbf{$\tau^2$-Noise and Vita-Noise.}
    To assess the robustness of agentic reasoning capability, we systematically analyze bias observed in real-world environments and design an automated noise injection pipeline that can inject realistic noise into arbitrary benchmarks.
    Based on this pipeline, we repeatedly and randomly inject noise into both the $\tau^2$ and Vita benchmarks to construct $\tau^2$-Noise and Vita-Noise, and report the averaged performance over multiple noisy instantiations.

    \item \textbf{Random Complex Tasks.}
    To evaluate the generalization of agentic reasoning capability, we introduce a new evaluation protocol, Random Complex Tasks, built upon an automated task synthesis process that randomly generates complex, executable, and verifiable agentic tasks across diverse scenarios inspired by our environment scaling pipeline.
    For each evaluation run, we randomly sample 100 diverse and complex tasks spanning more than four domains, and calculate the Avg@4 score. 
    To ensure reliability, we repeat the evaluation for three independent runs and report the average results across runs.
    
\end{itemize}
We will release the noise injection pipeline, as well as integrate Random Complex Tasks into an open evaluation platform, to support future reproducibility and benchmarking.

\paragraph{General QA}
We evaluate general reasoning on GPQA-Diamond~\citep{GPQA} and HLE~\citep{HLE}, which cover a broad range of knowledge-intensive and reasoning-focused tasks.
For GPQA-Diamond, we report Avg@16 to reduce variance introduced by sampling stochasticity.
For HLE, since our model is text-only, we report results on the text-only subset and ensure that all compared models follow the same setting.
We observe that the results of HLE are sensitive to prompt templates and scoring models.
To ensure fair comparison, all models are evaluated using the official HLE-recommended prompt template and are scored with the official o3-mini scoring model and template.

\paragraph{Coding}
We evaluate coding capability in two settings: code reasoning and agentic coding.
For code reasoning, we use LiveCodeBench~\citep{LiveCodeBench}, OJBench~\citep{OJBench}, and OIBench~\citep{OIBench}.
For LiveCodeBench, we evaluate on the 2408–2505 subset, covering the two most recent releases and comprising 454 problems.
We report Avg@4 for LiveCodeBench.
For OJBench and OIBench, due to evaluation cost and the long reasoning traces required by thinking models, we report Pass@1.
For agentic coding, we use SWE-bench Verified, which is a standard benchmark for software engineering agents.
We adopt the third-party agent framework R2E-Gym\footnote{\url{https://github.com/R2E-Gym/R2E-Gym}} as the execution backbone.
To ensure correctness, we manually clean and fix a small number of Docker images in the original benchmark, primarily due to dependency mismatches introduced by library upgrades, guaranteeing that all gold patches are executable.

\subsection{Main Results}
As illustrated in Table~\ref{tab:main_results}, we compare \longcat with several advanced open-weight and closed-weight reasoning models. 
Specifically, the open-weight models include DeepSeek-V3.2-Thinking~\citep{DeepSeek-V3.2}, Kimi-K2-Thinking~\citep{Kimi_K2_web_doc}, Qwen3-235B-A22B-Thinking-2507~\citep{Qwen3}, and GLM-4.7-Thinking~\citep{GLM-4.7}, while the closed-weight models include Claude-Opus-4.5-Thinking~\citep{Claude-Opus-4.5}, Gemini-3-Pro~\citep{Gemini3}, and GPT-5.2-Thinking-xhigh~\citep{GPT-5.2}.
Across a comprehensive suite of benchmarks, \longcat achieves highly competitive performance on traditional reasoning tasks and demonstrates strong advantages in agentic reasoning capabilities.
Unless otherwise specified, inference is conducted with temperature$=1.0$, top-$k=-1$, and top-$p=1.0$.
A detailed breakdown of these capabilities is provided in the subsequent analysis.

\begin{itemize}
\item \textbf{Mathematical Reasoning:} 
On challenging mathematical reasoning benchmarks, \longcat exhibits strong tool-integrated reasoning capability and consistently achieves first-tier performance.
In particular, when equipped with heavy mode, \longcat reaches performance comparable to leading closed-source models.
Specifically, \longcat with heavy mode attains a perfect score on AIME-2025, achieves a state-of-the-art score of 86.8 on IMO-AnswerBench, and delivers open-source state-of-the-art results on AMO-Bench.
Notably, although slightly behind the strongest closed-source models on AMO-Bench (EN), \longcat remains the best-performing open-source model.
Moreover, \longcat exhibits comparable performances on both the English and Chinese version of AMO-Bench, indicating advanced mathematical reasoning and tool-use capability in non-English settings.
\item \textbf{Agentic Search:} 
\longcat achieves state-of-the-art performance on both BrowseComp and BrowseComp-ZH.
With context management enabled, it attains 73.1 on BrowseComp and 77.7 on BrowseComp-ZH, surpassing all evaluated models.
On RWSearch, a private benchmark designed to evaluate real-world complex search scenarios, \longcat achieves a score of 79.5, second only to GPT-5.2-Thinking.

\item \textbf{Agentic Tool Using:} 
\longcat demonstrates state-of-the-art agentic tool-use capability among open-source models.
It achieves strong performance on $\tau^2$-Bench and VitaBench, including competitive results on their noise-augmented variants.
In particular, our model achieves state-of-the-art results on random complex tasks with arbitrarily generated tools.
These results indicate strong robustness to real-world environmental noise and excellent generalization to previously unseen task distributions.

\item \textbf{General QA:} 
\longcat maintains strong performance on general QA benchmarks while scaling agentic and tool-integrated reasoning.
It achieves a score of 25.2 on the text-only subset of HLE and a score of 85.2 on GPQA-Diamond under heavy mode, approaching open-source state-of-the-art results.

\item \textbf{Coding:} 
\longcat demonstrates competitive performance across both code reasoning and agentic coding benchmarks.
On algorithmic problem-solving tasks from the LiveCodeBench series, it ranks among the top open-source models.
On more difficult benchmarks such as OJBench and OIBench, our model achieves open-source second-best and best performance, respectively.
Notably, compared to GLM-4.7, \longcat achieves similar performance with substantially lower inference cost, requiring approximately 45k tokens per problem versus 57k tokens.
On SWE-bench Verified, \longcat performs competitively within the top tier of open-source models, further validating its capability in real-world software engineering tasks.

\end{itemize}

\section{One More Thing: Zig-Zag Attention Design}
Long-context efficiency has become an increasingly critical challenge for modern large language models. The steadily growing trend of longer reasoning traces poses a fundamental limitation to standard full attention, whose quadratic complexity quickly becomes prohibitively expensive for long-context agentic training and inference. Moreover, in heavy-thinking mode, inference latency is further amplified due to the simultaneous decoding of multiple parallel reasoning traces, making efficient attention mechanisms even more indispensable.
Existing approaches, including sparse and linear attention methods~\citep{sparse1, sparse2, Linear1, linear2}, attempt to mitigate this issue by reducing the computational complexity of attention. However, these methods typically require substantial retraining to adapt the model to a new attention architecture, which introduces considerable additional compute overhead and engineering cost.
To address this limitation, we explore an experimental efficient attention design and concurrently release an open-source model, LongCat-Flash-Thinking-ZigZag\footnote{\url{https://huggingface.co/meituan-longcat/LongCat-Flash-Thinking-ZigZag}}. Specifically, we propose Zigzag Attention, a sparse attention mechanism that enables existing full-attention models to be efficiently converted into sparse variants during mid-training. This conversion incurs only negligible overhead, while allowing the model to efficiently scale to ultra-long contexts, supporting sequence lengths of up to 1M tokens. 

Concretely, Zigzag Attention~\citep{zhang2026} combines Multi-head Latent Attention (MLA) with Streaming Sparse Attention (SSA)~\citep{DBLP:journals/corr/abs-2410-10819} to achieve computation that scales sub-quadratically with the full context length.
For each query token $h_t$, attention is restricted to a fixed set of key-value tokens consisting of (i) a local window of recent tokens and (ii) a small set of initial tokens at the beginning of the sequence.
Formally, the attention output is computed as:
\begin{equation}
u_t = \mathrm{Attn}\!\left(
h_t,\;
\{h_s \mid s \in [t-W, t] \cup [0, B)\}
\right),
\end{equation}
where $W$ denotes the local context window size and $B$ denotes the number of preserved prefix tokens.
Compared to full attention, this design significantly reduces computational and memory complexity while retaining both short-term context and global anchors.

\begin{figure}[ht]
    \centering
    \begin{subfigure}[b]{0.37\textwidth}
        \centering
        \includegraphics[width=\textwidth]{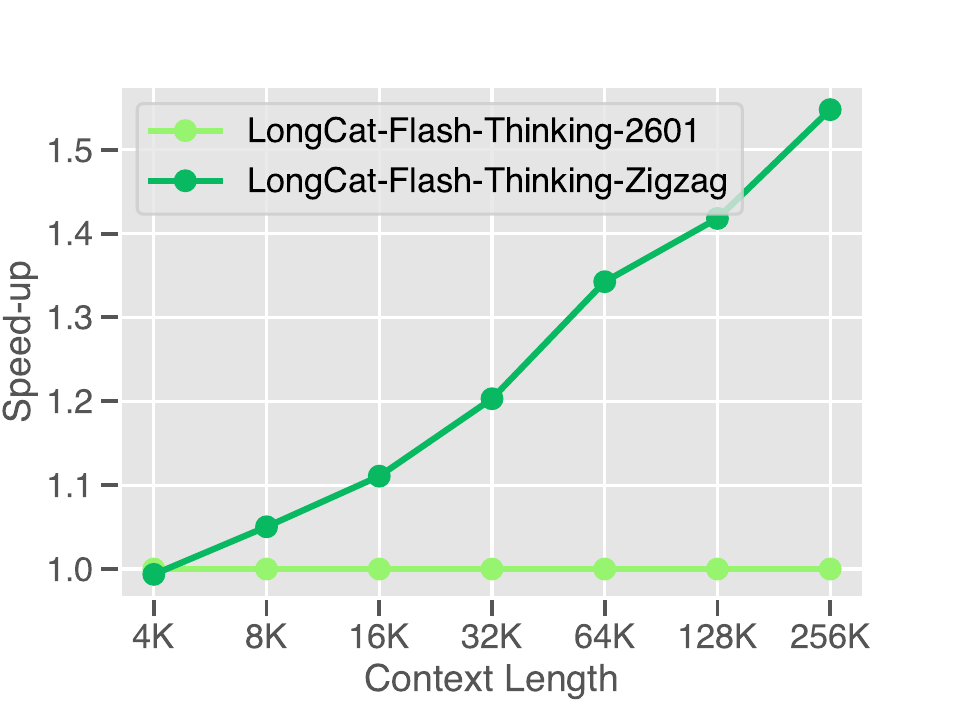}
        \caption{Prefill.}
        \label{fig:efficiency_prefill}
    \end{subfigure}
    % \hfill
    \begin{subfigure}[b]{0.37\textwidth}
        \centering
        \includegraphics[width=\textwidth]{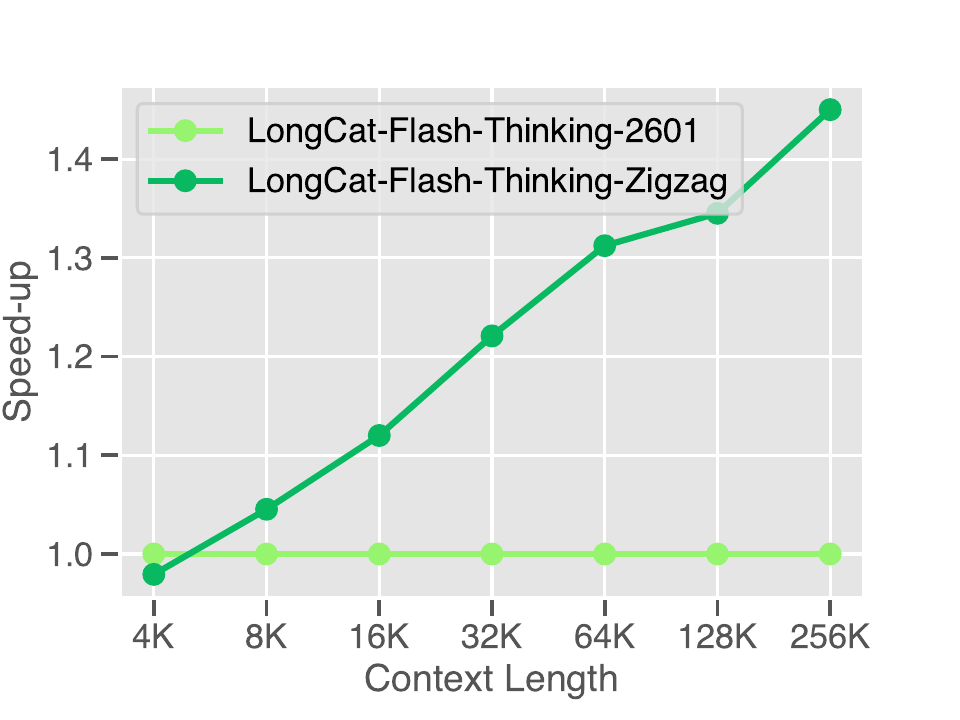}
        \caption{Decode.}
        \label{fig:efficiency_decode}
    \end{subfigure}
    \caption{Inference efficiency comparison between LongCat-Flash-Thinking and LongCat-Flash-Thinking-ZigZag with Zigzag Attention. Speedups are measured on production inference clusters.}
    \label{fig:zigzag_efficiency}
\end{figure}

\paragraph{Zigzag Conectivity}
Zigzag Attention adopts a layer-wise interleaved sparsification strategy.
Specifically, approximately 50\% of full-attention layers are replaced with SSA layers, while the remaining layers retain MLA-based full attention.
This layer-level sparsity avoids the computational imbalance and GPU thread divergence commonly introduced by head-level sparsification, leading to more efficient hardware utilization.
Although attention within each SSA layer is sparse and local, global information is preserved through cross-layer composition.
By alternating sparse and full-attention layers, information propagates across distant positions over multiple layers, forming a zigzag-shaped connectivity path along the sequence.
As a result, long-range dependencies remain accessible despite per-layer sparsity.

\paragraph{Zigzag Integration}
Zigzag Attention is introduced during mid-training through a structured sparsification procedure.
First, we use a calibrated dataset to estimate the relative importance of attention layers in the pretrained model.
Second, the subset of layers with the lowest importance scores is replaced with SSA layers.
After sparsification, the model undergoes continued long-context mid-training, together with YaRN-based positional encoding extension, enabling context lengths of up to 1M tokens.
In practice, we adopt a block size of 128, with one sink block and seven local blocks, resulting in an effective attention span of 1,024 tokens per layer.
Replacing approximately half of the full-attention layers with Zigzag Attention yields about a $1.5\times$ end-to-end inference speedup, as shown in Figure~\ref{fig:zigzag_efficiency}, while preserving reasoning performance and agentic capabilities across benchmarks.

To unlock the ability of handling a longer context, we equip these recipes with YaRN~\citep{DBLP:conf/iclr/PengQFS24} so that LongCat-Flash-Thinking-ZigZag can extrapolate itself to processing up to 1M tokens. In addition to that, we provide a few crucial parameters involved in the model. The block size is 128, the number of sink blocks is 1, and the number of local blocks is 7, summing to 1,024 tokens. 

\begin{figure}[ht]
    \centering
    \includegraphics[width=\linewidth]{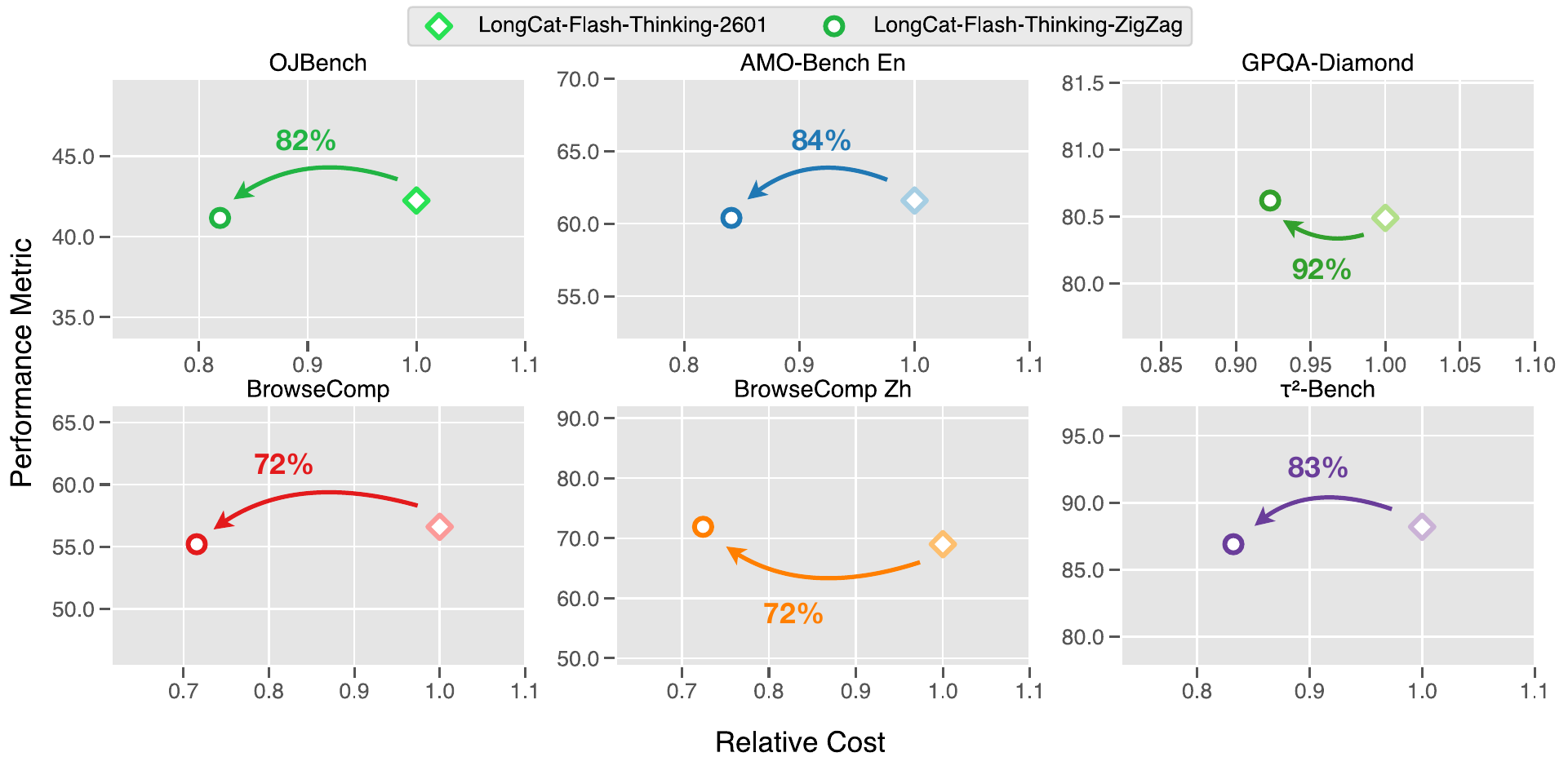}
    \caption{The performance versus relative cost. The percentages attached to arrows indicate the reduced cost when benchmarking the performance on the concerned datasets.}
    \label{fig:zigzag_pareto}
\end{figure}

LongCat-Flash-Thinking-ZigZag yields a good trade-off between performance and speed. A quick glimpse on how it improves efficiency yet maintaining competitive performance is illustrated in Figure~\ref{fig:zigzag_pareto}.

\section{Conclusion}
We introduce \longcat, an open-weight 560B MoE reasoning model with superior agentic reasoning capability.
This model is built upon a unified, end-to-end training pipeline that co-designs architecture, pre-training, post-training, data construction, and scalable RL infrastructure, all explicitly tailored to support long-horizon, interaction-driven agentic reasoning.
The core innovations underpining \longcat are as follows: 
(i) A scalable environment construction and multi-domain reinforcement learning framework that enables stable acquisition of generalizable agentic skills.
(ii) A robust agentic training pipeline that systematically incorporates real-world environmental noise through curriculum-based reinforcement learning.
(iii) A Heavy Thinking Mode that enables effective test-time scaling of reasoning by jointly expanding reasoning width and depth.
Through this holistic design, \longcat sets new state-of-the-art performance among open-source models across a wide range of agentic benchmark, narrowing the performance gap with leading closed-source models.

\newpage
\section*{Contributions}

The listing of authors is in alphabetical order.
Entries with identical English names will be sorted based on the order of Chinese stroke count and pronunciation. An asterisk (*) indicates members who have departed from the team.

\begin{CJK}{UTF8}{gbsn}

\begin{center}

\begin{tabular}{p{0.185\textwidth}p{0.185\textwidth}p{0.185\textwidth}p{0.185\textwidth}p{0.185\textwidth}}
Anchun Gui & Hongyan Hao & Mingyang Zhu & Xiandi Ma & Yongwei Zhou \\
Bei Li & Hongyin Tang & Peiguang Li & Xiangcheng Liu & Youshao Xiao \\
Bingyang Tao & Hongyu Zang* & Peng Pei & Xiangyu Xi & Yu Wang \\
Bole Zhou & Hongzhi Ni & Peng Zhao & Xiangyuan Liu & Yu Yang \\
Borun Chen & Hui Su & Pengcheng Jia & Xiangzhou Huang & Yuchen Xie \\
Chao Zhang(张超) & Jiacheng Zhang & Pengtao Zhang & Xiao Liu & Yuchen Yu \\
Chao Zhang(张朝) & Jiahong Zhou & Ping Liu & Xiaodong Cai & Yuchuan Dai \\
Chen Gao & Jiahuan Li & Qi Gu & Xiaolong Chen & Yue Xu \\
Chen Zhang & Jiaming Wang & Qiong Huang & Xiaowei Shi & Yueqing Sun \\
Chengcheng Han & Jian Yang & Qiyuan Duan & Xiaoyu Li & Yufei Zhang \\
Chenhui Yang & Jianfei Zhang & Quanchi Weng & Xin Chen & Yuhuai Wei \\
Chuyu Zhang & Jianhao Xu & Rongxiang Weng & Xingchen Liu & Yulei Qian \\
Cong Chen & Jianing Wang & Rongzhi Zhang & Xuan Huang & Yunfan Liang \\
Cunguang Wang & Jiapeng Zhu & Rumei Li & Xuezhi Cao & Yunke Zhao \\
Daoru Pan & Jiaqi Sun & Shanglin Lei & Xunliang Cai & Yuwei Jiang \\
Defei Bu & Jiarong Shi & Shengnan An & Yan Chen & Yuxin Bian \\
Dengchang Zhao & Jiarui Zhao & Shijun Dai & Shizhe Wu & Yang Bai \\
Yuxin Chen & Di Xiu & Jingang Wang & Shuaikang Liu & Yang Liu \\
Yuxin Liu & Dishan Liu & Jinluan Yang & Shuang Zhou & Yang Yang \\
Zeyang Yu & Dongyu Ru & Jinrui Ding & Shuo Wang & Yang Zheng \\
Zhao Yang & Dunwei Tu & Jinwei Xiao & Songyuan Zhao & Yanyu Chen \\
Zhengsheng Huang & Fan Wu & Jiyuan He & Tao Liang & Yaoming Wang \\
Zhengyu Chen & Fengcheng Yuan & Juncan Xu & Tianhao Hu & Yaoming Zhu \\
Zhijian Liu & Fengcun Li & Kefeng Zhang & Tianze Chen & Yaorui Shi \\
Zhikang Xia & Gang Xu & Keheng Wang & Wei Liu & Yaqi Huo \\
Zhimin Lin & Guanyu Wu & Li Wei & Wei Shi & Yerui Sun \\
Zhiyuan Yao & Guoyuan Lin & Lianhui Ma & Wei Wang & Yi Zhang \\
Zhuofan Chen & Haibin Wang & Lin Qiu & Weifeng Tang & Yifan Lu \\
Zhuowen Han & Hansi Yang & Lingbing Kong & Wenjie Shi & Yifan Zhao \\
Zijian Zhang & Hao Yang & Lingchuan Liu & Wenlong Zhu & Yihao Chen \\
Ziran Li & Haonan Yan & Linsen Guo & Wentao Chen & Yi-Kai Zhang \\
Ziwen Wang & Haoxiang Ma & Mengshen Zhu & Wentao Shi* & Yitao Zhai \\
Ziyuan Zhuang & Haoxing Wen & Mengxia Shen & Xi Su & Yongjing Yin \\
LongCat-Flash \\
\end{tabular}

\end{center}

\end{CJK}

\bibliographystyle{unsrtnat}
\bibliography{references}

\newpage
\appendix

\section{Optimal Hyperparameter Prediction}
\label{sec:optimal_hyperparameter_prediction}

\begin{figure}[t]
    \centering
    \includegraphics[width=0.8\textwidth]{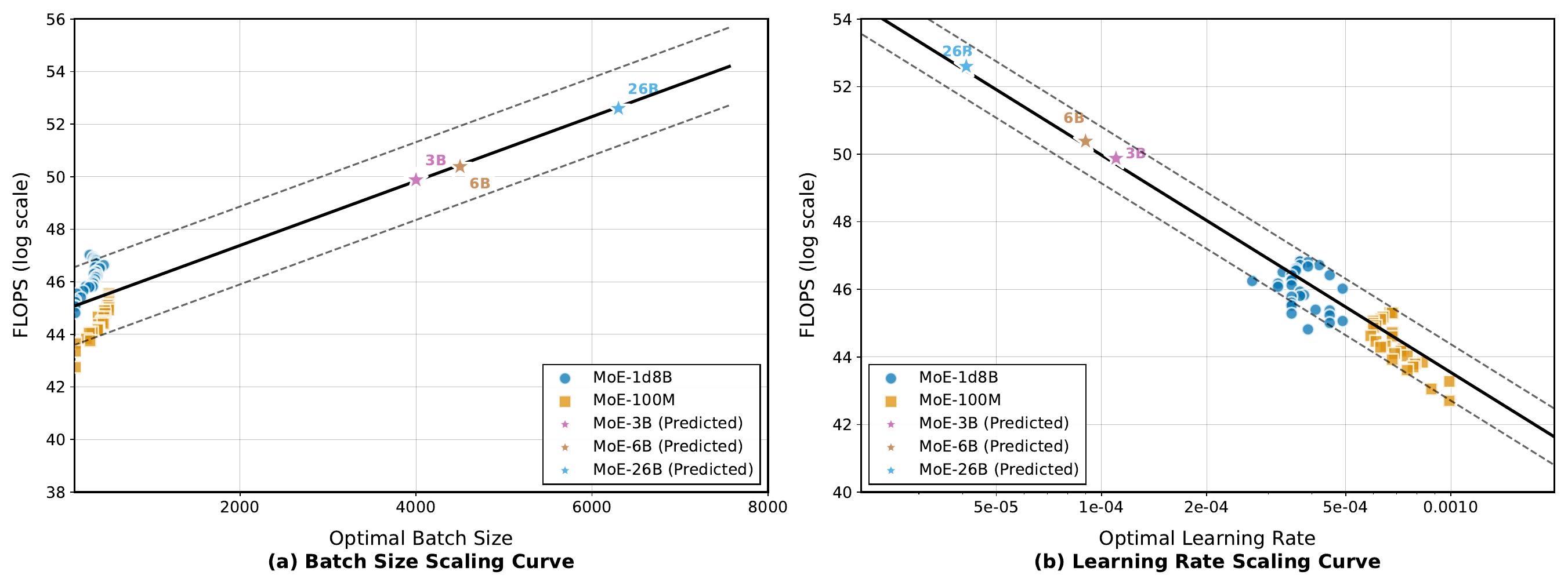}
    \caption{Scaling curves of optimal batch size and learning rate. The circles and squares represent MoE-1.8B and MoE-100M models respectively. The dashed lines show the power-law fitting with confidence intervals. The scaling laws are first validated on MoE-3B and MoE-6B models, then extrapolated to predict optimal configurations for MoE-26B.}
\label{fig: hyparam_scaling}
\end{figure}

Efficiently identifying the optimal hyperparameters is one of the core challenges in large-scale mid-training, given the vast search space and high computational costs.  
To address this challenge, we propose a novel method for predicting optimal hyperparameters, specifically designed to minimize the computational cost of finding the best configuration.

Our approach consists of two key steps:
\begin{itemize}
    \item \textbf{Hyperparameter Mapping}:  
    We train small models with different hyperparameters, using validation loss and FLOPS to map the optimal hyperparameters to their computational cost (see Figure \ref{fig: hyparam_scaling}), providing insights into how configurations affect training efficiency and performance.
    
    \item \textbf{Hyperparameter Prediction}:  
    For a given continual training checkpoint, we estimate the equivalent compute cost using validation loss, which means the amount of compute required to achieve the same loss when training from scratch on the continual training data. We then predict the optimal hyperparameters based on this estimate and the actual computational load.
\end{itemize}

Through this approach, we can predict optimal hyperparameters that enable efficient continual training, improving model performance with minimal computational overhead.

\section{Token Threshold Context Management Performance Evaluation}
\label{sec:token_threshold_performance_evaluation}

\begin{figure}[t]
    \centering
    \includegraphics[width=0.45\textwidth]{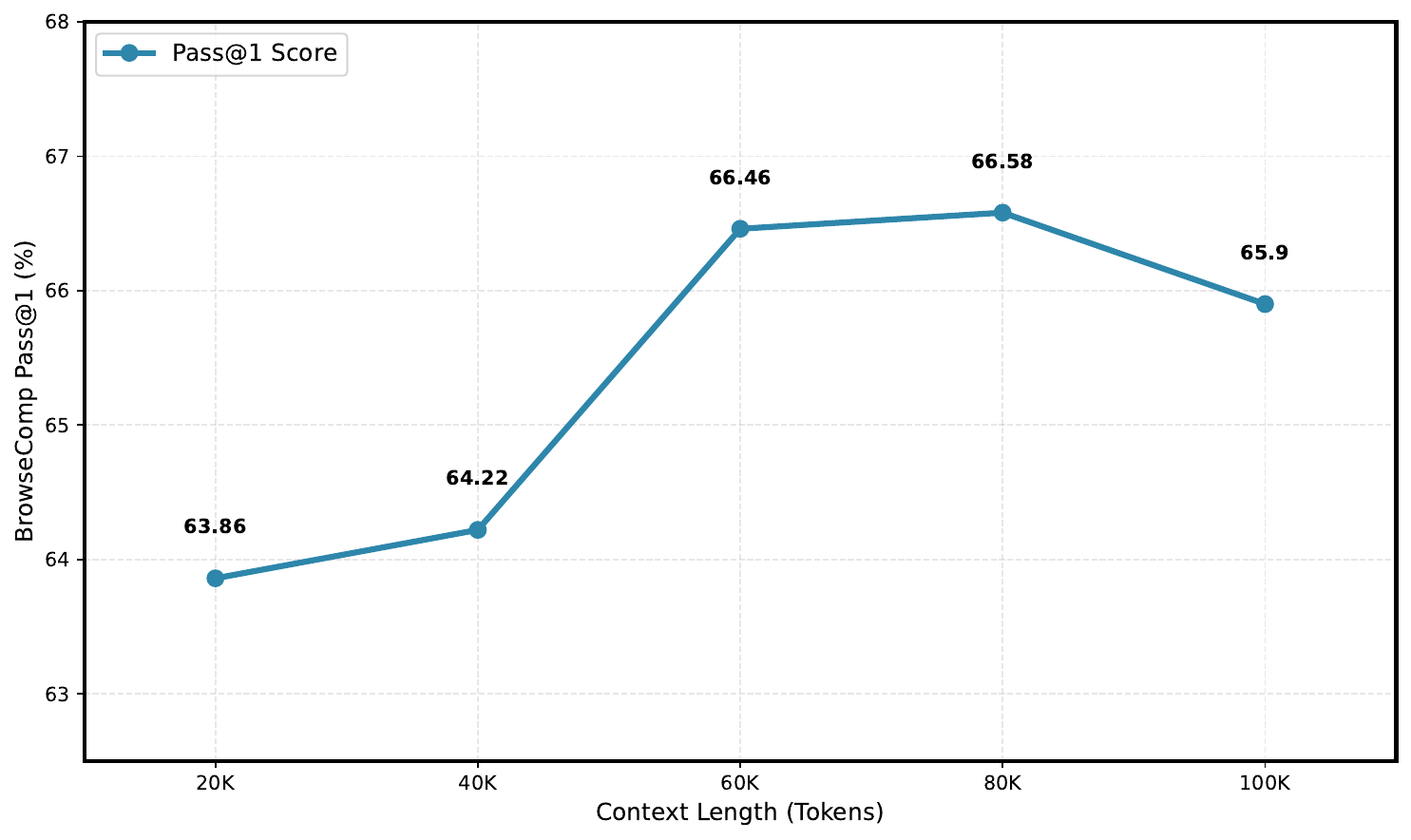}
    \caption{Pass@1 accuracy of BrowseComp with varying summary context token lengths.}
    \label{fig:context_management_summary_figure}
\end{figure}

As shown in Figure \ref{fig:context_management_summary_figure}, we evaluate BrowseComp’s Pass@1 accuracy across varying summary context token lengths, with a maximum context turn limit of 500. Accuracy rises steadily from 63.86\% at 20K tokens to a peak of 66.58\% at 80K tokens, then drops to 65.9\% at 100K tokens. This identifies 80K as the optimal context length for summary-based context management, so we fix 80K context length as the summarization trigger threshold in all subsequent experiments.

\end{document}